\documentclass[10pt,twocolumn,letterpaper]{article}

\usepackage{cvpr}              % To produce the CAMERA-READY version
\usepackage{graphicx}
\usepackage{amsmath}
\usepackage{amssymb}
\usepackage{booktabs}
\usepackage{subcaption}
\usepackage{xcolor}
\usepackage{multirow}
\usepackage{colortbl}
\usepackage{natbib}
\usepackage{float}

\usepackage[pagebackref,breaklinks,colorlinks]{hyperref}

\usepackage[capitalize]{cleveref}
\crefname{section}{Sec.}{Secs.}
\Crefname{section}{Section}{Sections}
\Crefname{table}{Table}{Tables}
\crefname{table}{Tab.}{Tabs.}

\def\paperID{15848}
\def\confName{CVPR}
\def\confYear{2025}

\renewcommand\paragraph[1]{\noindent\textbf{#1}}

\begin{document}

\title{Evaluating Model Perception of Color Illusions in Photorealistic Scenes}

\author{%
Lingjun Mao \hspace{1cm} Zineng Tang \hspace{1cm} Alane Suhr \\
  University of California, Berkeley \\
  \texttt{\{lingjun, terran, suhr\}@berkeley.edu} \\
}

\maketitle

%%%%%%%%% BODY TEXT
\begin{abstract}

\begin{figure*}[htbp]
    \centering
    \begin{subfigure}{0.32\textwidth}
        \centering
        \includegraphics[width=\linewidth]{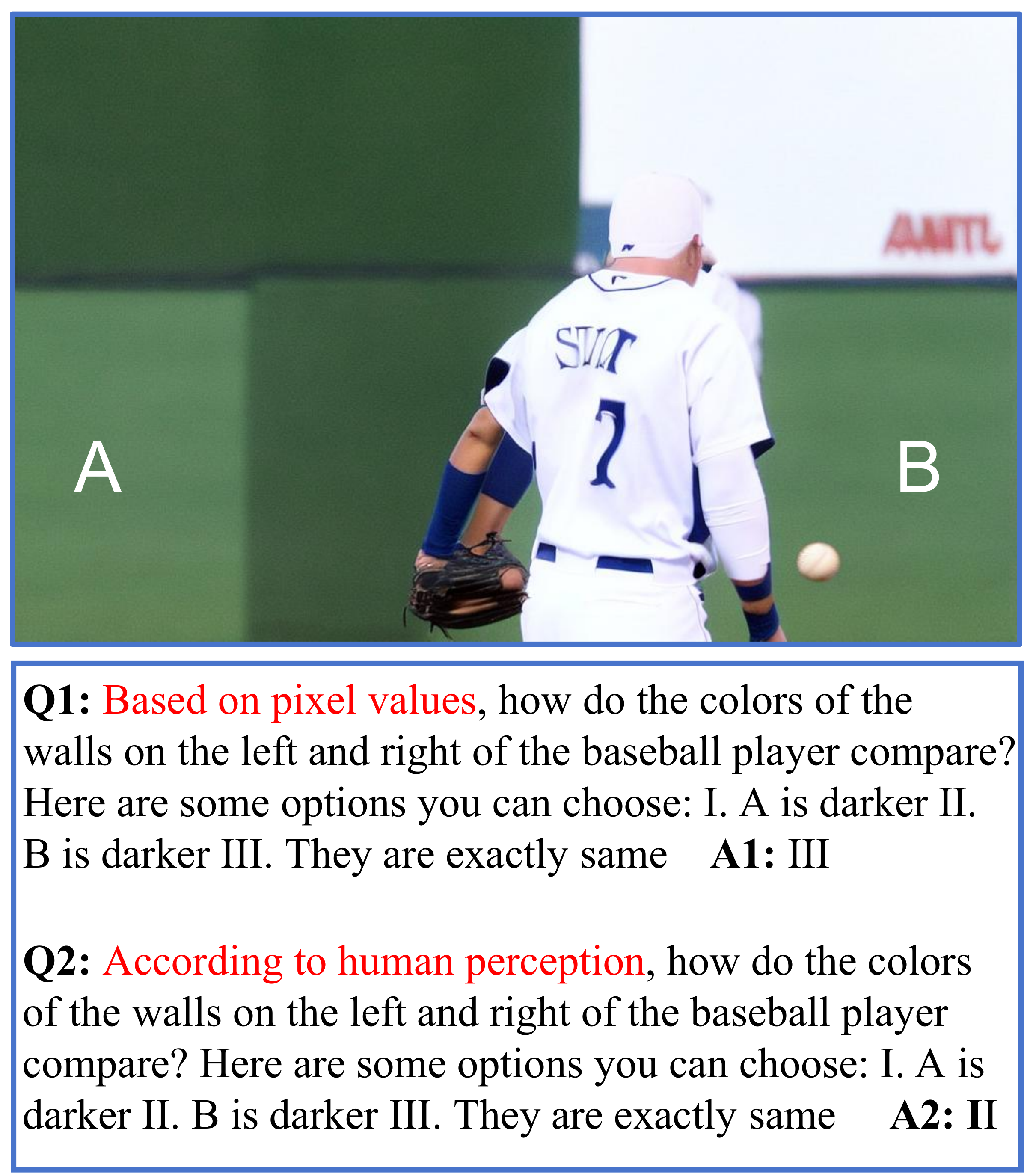}
        \caption{Contrast Illusion}
        \label{fig:subfig1}
    \end{subfigure}
    \hfill
    \begin{subfigure}{0.32\textwidth}
        \centering
        \includegraphics[width=\linewidth]{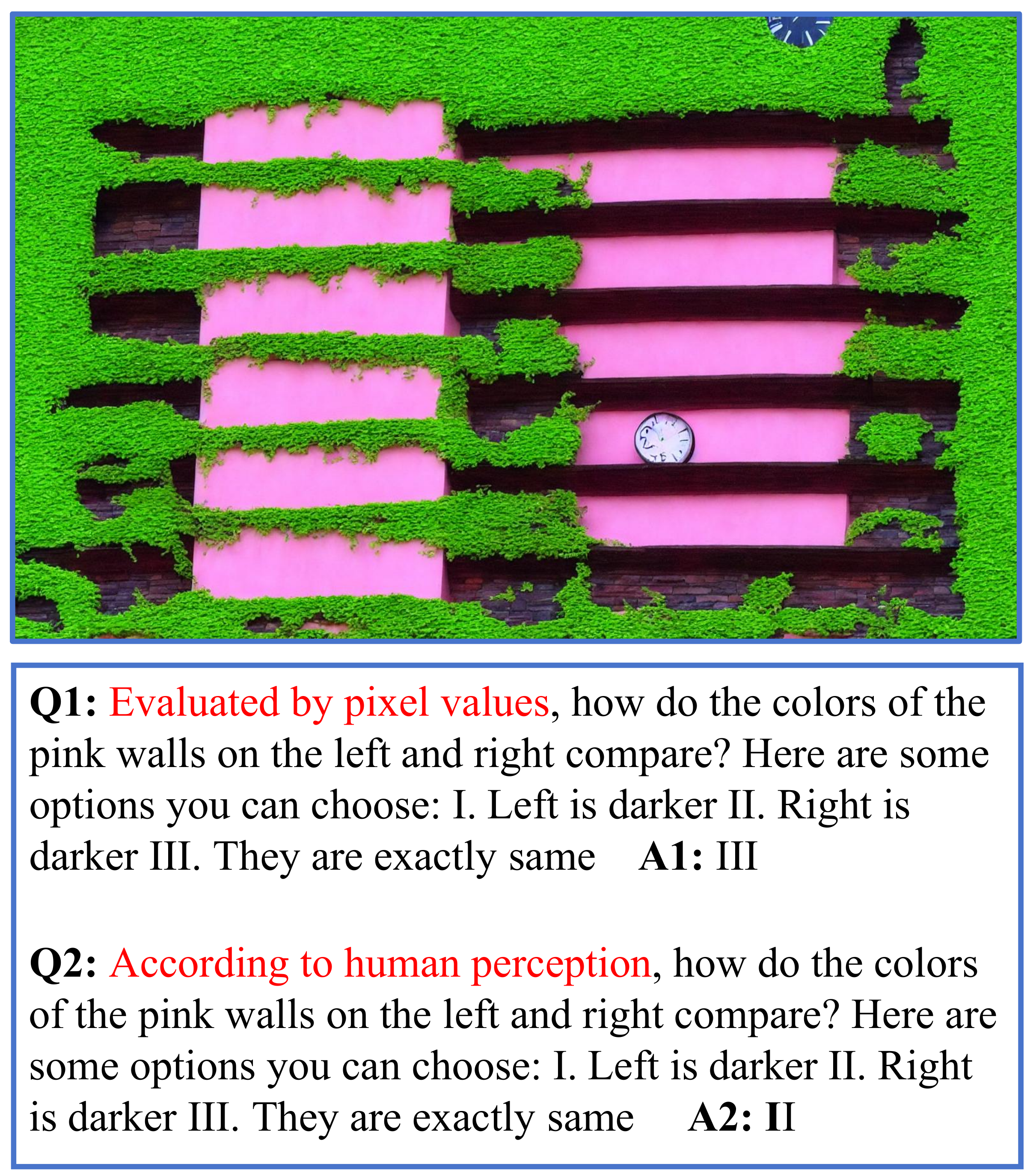}
        \caption{Stripe Illusion}
        \label{fig:subfig2}
    \end{subfigure}
    \hfill
    \begin{subfigure}{0.32\textwidth}
        \centering
        \includegraphics[width=\linewidth]{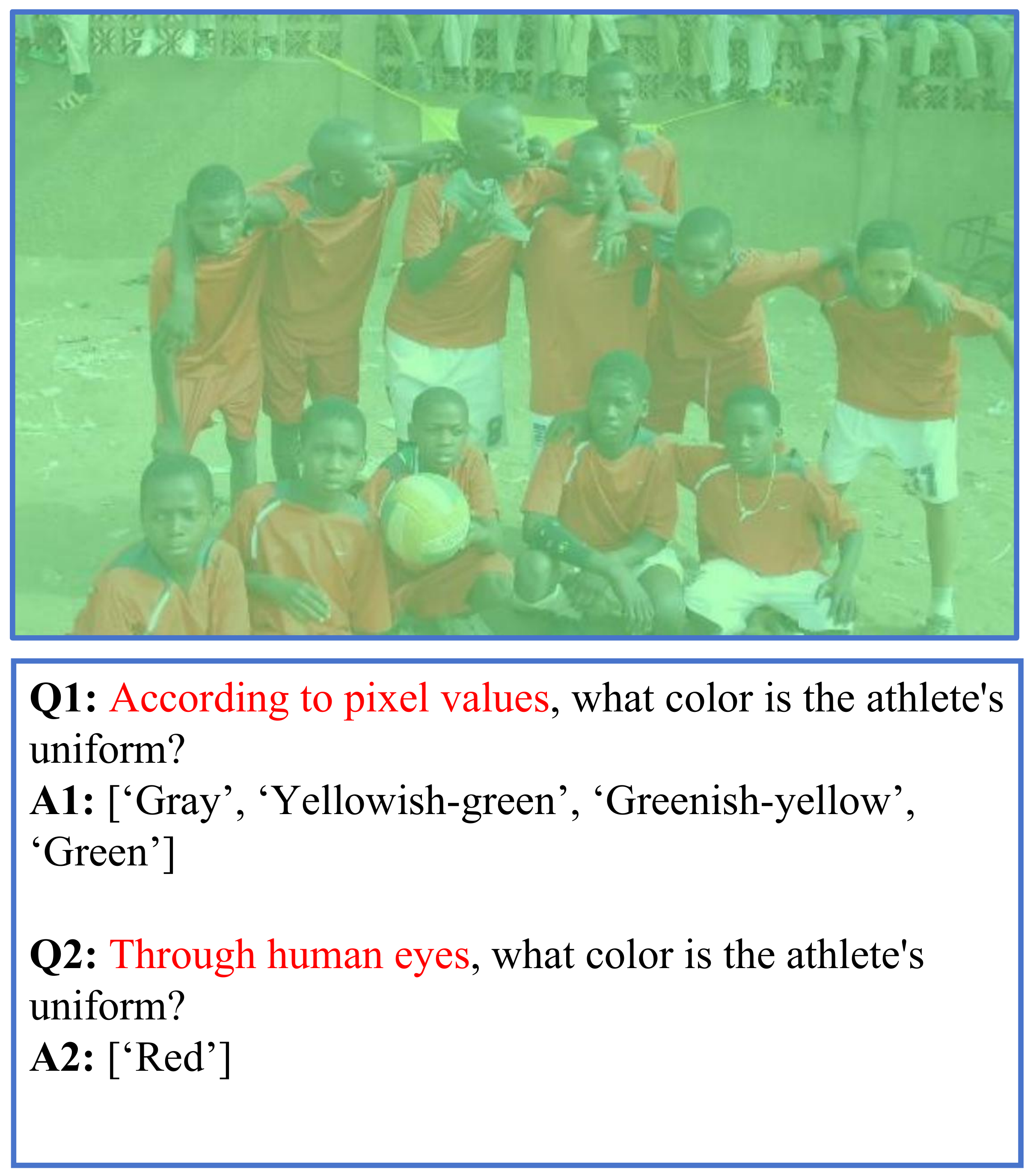}
        \caption{Filter Illusion}
        \label{fig:subfig3}
    \end{subfigure}
    \caption{\textbf{Examples of color illusions.} (a) Contrast illusion: The two green squares on the left and right are identical in color, but most people perceive the square on the right as darker due to the brighter background. (b) Stripe illusion: The left and right walls seem to be different colors, but this is an illusion caused by the dark stripes interfering with our perception. (c) Filter illusion: The player's uniform appears red, but in reality, there are no red pixels present.}
    \label{fig:illusion_examples}
\end{figure*}

We study the perception of color illusions by vision-language models. Color illusion, where a person's visual system perceives color differently from actual color, is well-studied in human vision. However, it remains underexplored whether vision-language models (VLMs), trained on large-scale human data, exhibit similar perceptual biases when confronted with such color illusions. We propose an automated framework for generating color illusion images, resulting in RCID (Realistic Color Illusion Dataset), a dataset of 19,000 realistic illusion images. Our experiments show that all studied VLMs exhibit perceptual biases similar human vision. Finally, we train a model to distinguish both human perception and actual pixel differences. 
% Our code and dataset can be found in \url{https://Color-Illusion.github.io}.
\end{abstract}

\section{Introduction}

Human visual perception does not always align perfectly with the real world~\citep{gregory1968visual, carbon2014understanding, gentilucci1996visual}; we are easily influenced by contextual factors such as lighting conditions or surrounding objects and patterns. Figure~\ref{fig:illusion_examples} shows several examples of color illusions, where human judgment of object color and brightness does not align with actual pixel values. 
% This phenomenon is referred to as \textbf{color illusion}. 
Currently, many large vision-language models (VLMs)~\citep{liu2023visualinstructiontuning, liu2024improvedbaselinesvisualinstruction, dai2023instructblipgeneralpurposevisionlanguagemodels,wang2024cogvlmvisualexpertpretrained} exhibit behavioral similaritities to human visual perception in various aspects, including object recognition~\citep{zhang2024visionlanguagemodelsvisiontasks}, scene understanding~\citep{10711845, cui2024open}, and spatial reasoning~\citep{chen2024spatialvlmendowingvisionlanguagemodels, tang2024groundinglanguagemultiperspectivereferential}. 
But because the human visual system behaves idiosyncratically when encountering these illusions, we aim to study whether these models perceive such illusions in the same way humans do.
Prior work~\citep{shahgir2024illusionvqa, zhang2023grounding} has  explored this question by sourcing illusion images from the internet. However, this approach has a  drawback: the number of illusion images on the web is limited, and most images (e.g., 60\% for the IllusionVQA dataset~\citep{shahgir2024illusionvqa}) are well-known examples of these illusions; thus, VLMs have likely memorized humanlike behavioral responses to them. 
% as the models may have already seen these images and know the correct answers in advance, rather than answering with their own direct observation. 
% A typical example of this is shown in Figure~\ref{fig:chat_examples}. 
Additionally, the limited scale also restricts the depth and variety of analyses that can be conducted.

To address these limitations, we propose an automatic framework for generating realistic color illusion images, covering many scenarios with illusions that we might genuinely encounter in our real life. We cover three different types of color illusions: \textit{contrast} (Figure~\ref{fig:illusion_examples}a), \textit{stripe} (Figure~\ref{fig:illusion_examples}b), and \textit{filter} (Figure~\ref{fig:illusion_examples}c) illusions. For contrast and stripe illusions, we first generate simple arrangements of shapes exhibiting specific illusions, which are then processed through ControlNet~\citep{zhang2023addingconditionalcontroltexttoimage} to generate more realistic images. For filter illusions, we directly apply filters to the original MS COCO 2017 images. %to create illusion effects.
In the end, we produce a dataset consisting of total 19,000 images, half of which include an illusion, and the other half being a control set without illusions. 
Each image is paired with a synthetically-generated natural language question, asking either about color differences between specific regions or directly inquiring about the color of a particular object in the image.

We test recent VLMs using our generated illusion images. We find that, after appropriate fine-tuning, they perform well on non-illusion images, but their accuracy drops significantly on illusion images. Some of their responses exhibit visual biases similar to those of humans, while some are entirely incorrect, failing to align with either pixel values or human perception. We also observe that external prompts, such as Chain of Thought (CoT) \citep{wei2023chainofthoughtpromptingelicitsreasoning} and few-shot examples, can influence VLM performance to some extent. Furthermore, fine-tuning on illusion images enhances the models' ability to understand these color illusions. Based on this, we propose a foundational baseline method that enables models to capture both human perceptual biases and true pixel values. Finally, we investigate the underlying causes of VLM susceptibility to color illusions, including biases from both the visual system and prior knowledge.

% \begin{figure}[htbp]
%     \centering
%     \begin{subfigure}{0.48\textwidth}
%         \centering
%         \includegraphics[width=\linewidth]{imgs/chat_example1.pdf}
%         \caption{Ask GPT-4 using an online-sourced image}
%         \label{fig:ask_using_online_image}
%     \end{subfigure}
%     \hfill
%     \begin{subfigure}{0.48\textwidth}
%         \centering
%         \includegraphics[width=\linewidth]{imgs/chat_example2.pdf}
%         \caption{Ask GPT-4 using a generated image}
%         \label{fig:ask_using_generated_image}
%     \end{subfigure}
%     \caption{Both images depict the same illusion, where identical shades of gray appear darker or lighter due to the background. When ask about the online image, GPT-4 accurately identify the illusion. However, with the generated image, GPT-4 is deceived and give an incorrect response.}
%     \label{fig:chat_examples}
% \end{figure}

In conclusion, our contributions are as follows: (1) We propose an automated framework for generating realistic illusion images and create a large, realistic dataset of color illusion images, named \textbf{RCID} (\underline{R}ealistic \underline{C}olor \underline{I}llusion \underline{D}ataset), to enhance the fairness and accuracy of model testing.
(2)  We investigate underlying mechanisms of color illusions in VLMs, highlighting the combined influence of the vision system and prior knowledge. We also explore how external prompts and instruction tuning impact the models' performance on these illusions.
(3) We propose an simple training method that enables models to understand human perception while also recognizing the actual pixel values.
Our code and data are released under an open-source license upon publication at the following URL:  \url{https://github.com/mao1207/RCID}. % upon publication.

% ICLR requires electronic submissions, processed by
% \url{https://openreview.net/}. See ICLR's website for more instructions.

% If your paper is ultimately accepted, the statement {\tt
%   {\textbackslash}iclrfinalcopy} should be inserted to adjust the
% format to the camera ready requirements.

% The format for the submissions is a variant of the NeurIPS format.
% Please read carefully the instructions below, and follow them
% faithfully.

\section{Related Work}

% \paragraph{Visual Illusions in Humans}
Visual illusion is a phenomenon where human perception differs from the actual environment~\citep{gregory1968visual, carbon2014understanding, gentilucci1996visual}, often triggered by specific visual stimuli~\citep{cheng2023reconstructing}. These illusions offer valuable insights into human perception~\citep{gregory1968perceptual, qiu2008muller} and have been widely utilized by artists in their works~\citep{livingstone1988art, gombrich2023art}. 
Many studies~\citep{maddaluno2018top, meyer1987top} have attempted to explore the causes of visual illusions, with the \textit{bottom-up} and \textit{top-down} theories being the most widely accepted explanations. Proponents~\citep{saeedi2024brightness} of the \textit{bottom-up} theory argue that low-level visual signals, such as contrast, brightness, edges, and motion, are correlated with human perception of illusions. However, supporters of the \textit{top-down} theory contend that prior knowledge is also a major factor in the occurrence of these illusions. Evidence shows that people in industrialized societies are more susceptible to the Müller-Lyer illusion due to being misled by implied arrow meanings, and individuals from language communities with fewer color terms are less affected by color illusions~\citep{alter2014drunk, phillips2019cross}.

% \paragraph{Visual Illusions in Machines}
Motivated by illusion phenomena in human visual perception, recent research has begun to explore whether models are also susceptible to such illusions. Early research~\citep{gomez2019convolutional, gomez2020color} compares the model's reconstruction of the original pixel input with human perception, suggesting that traditional vision models, such as CNNs, are influenced by visual illusions in a way similar to human perception. More recent studies~\citep{shahgir2024illusionvqa, zhang2023grounding} have shown that vision-language models also exhibit similar perceptual biases. However, images on which these models have been tested are either sourced from the internet or manual compositions of simple shapes (e.g., two squares of different colors), making it easy for the VLM to infer results based on their knowledge rather than actual observation. Additionally, the number of these images is quite limited.
To address these limitations, we develop a large-scale dataset that embeds visual illusions into realistic scenes, enabling robust evaluation in complex, real-world contexts. Additionally, our study fills a gap by investigating the mechanisms behind VLMs' susceptibility to illusions.

\begin{figure*}[htbp]
\centering
  \includegraphics[width=0.9\textwidth]{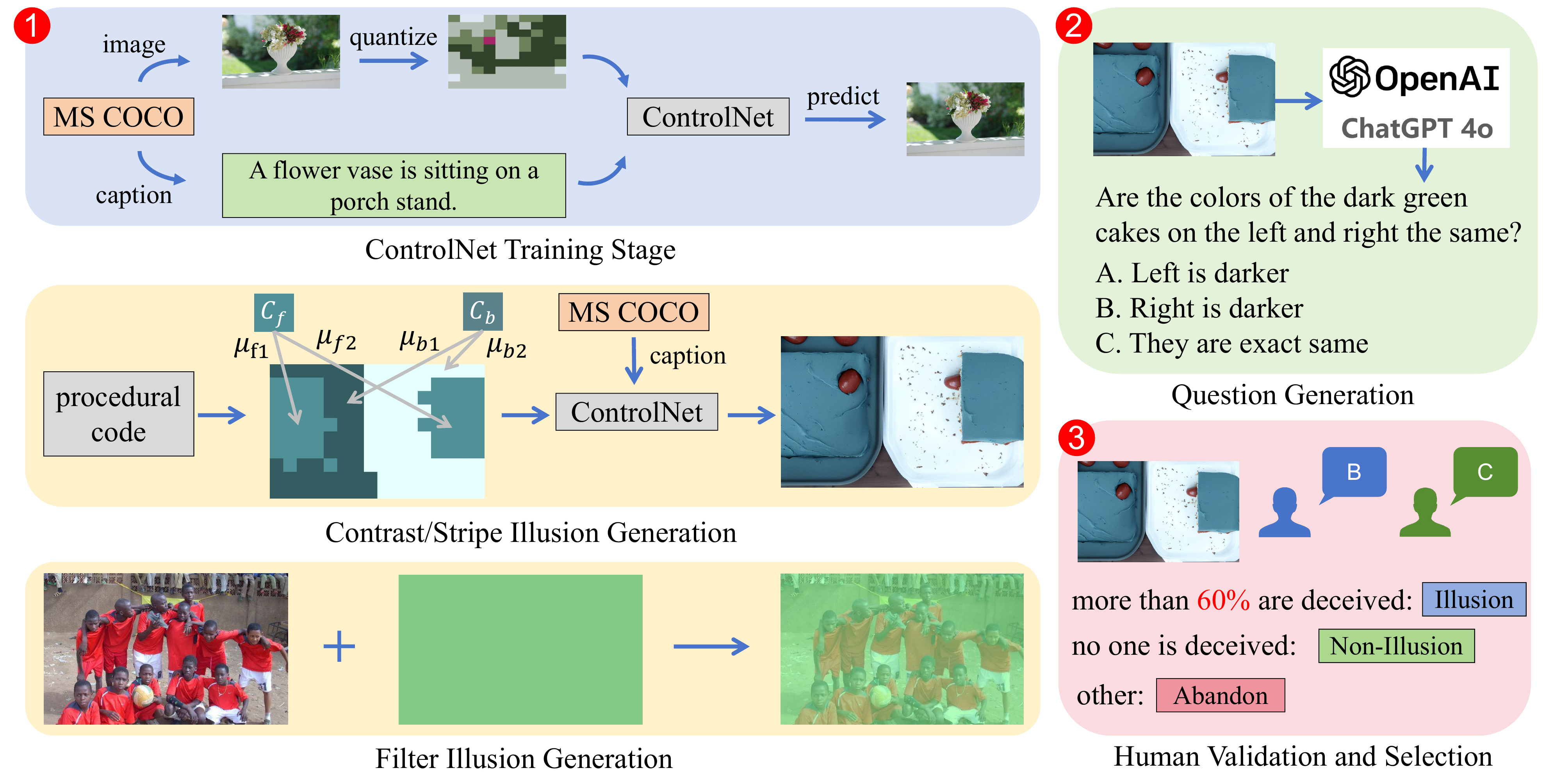}
    \caption{Process for generating our dataset.}
    % \vskip -1em
    \label{fig:framework_of_data_generation}
\end{figure*}

\vspace{-0.01cm}

\section{Illusion Dataset Construction}

The construction of our dataset involves three steps (Figure~\ref{fig:framework_of_data_generation}): (1) \textbf{Image Generation.} For contrast and stripe illusions, we use procedural code to generate simple illusion images, which are then processed by ControlNet to create realistic illusion images. For filter illusions, we directly apply contrasting color filters to the original images. Each type of illusion also includes a corresponding control group without any illusions for comparison. (2) \textbf{Question Generation.} We use GPT-4o to generate image-specific questions that  are designed to evaluate the model's understanding of the illusion. (3) \textbf{Human Feedback.} We collect human participants' feedback on these images and adjust the original classification of ``illusion'' and ``non-illusion'' based on whether participants are deceived. 
% The overall process is illustrated in Figure~\ref{fig:framework_of_data_generation}.

\subsection{Contrast and Stripe Illusion Image Generation}

\paragraph{ControlNet Training Stage:} We begin by training a ControlNet to map from  simple illusion images to  photorealistic images with illusions. We train using modified image-caption pairs \( (I, T) \) from the MS COCO 2017 dataset. First, we quantize each original image \( I \) into a \( 10 \times 10 \) grid \( G \), where each cell \( G_{x, y} \) represents the average color of its corresponding region:

\begin{small}
\begin{equation*}
    G_{ x, y} = \frac{1}{|R_{x, y}|} \sum_{p \in R_{x, y}} I(p)
\end{equation*}
\end{small}

\noindent Here, \(R_{x, y}\) denotes the set of pixels in the \((x, y)\)-th grid cell, and \(I(p)\) is the color value of pixel \(p\) in the original image.

During training, Gaussian noise is added to the quantized image \( x_0 = G \), producing a noisy image \( x_t \) at time step \( t \). The model, conditioned on the text prompt \( T \) and color distribution \( G \), is trained to predict the added noise \( \epsilon_\theta \) by minimizing the expected squared error:

\begin{small}
    \begin{equation*}
\mathcal{L} = \mathbb{E}_{x_0, t, C, \epsilon \sim \mathcal{N}(0, I)} \left[ \left\| \epsilon - \epsilon_\theta(x_t, t, T, G) \right\|_2^2 \right]
    \end{equation*}
\end{small}

Through this loss, the model learns to progressively refine \( x_t \) into a realistic image during inference, guided by color distribution and the text prompt.

% \begin{figure}[H]
% \centering
% \includegraphics[width=\linewidth]{imgs/procedural.pdf}
%     \caption{Procedural Generation Stage}
%     % \vskip -1em
%     \label{fig:location}
% \end{figure}

\begin{figure*}[t]
\centering
  \includegraphics[width=0.9\textwidth]{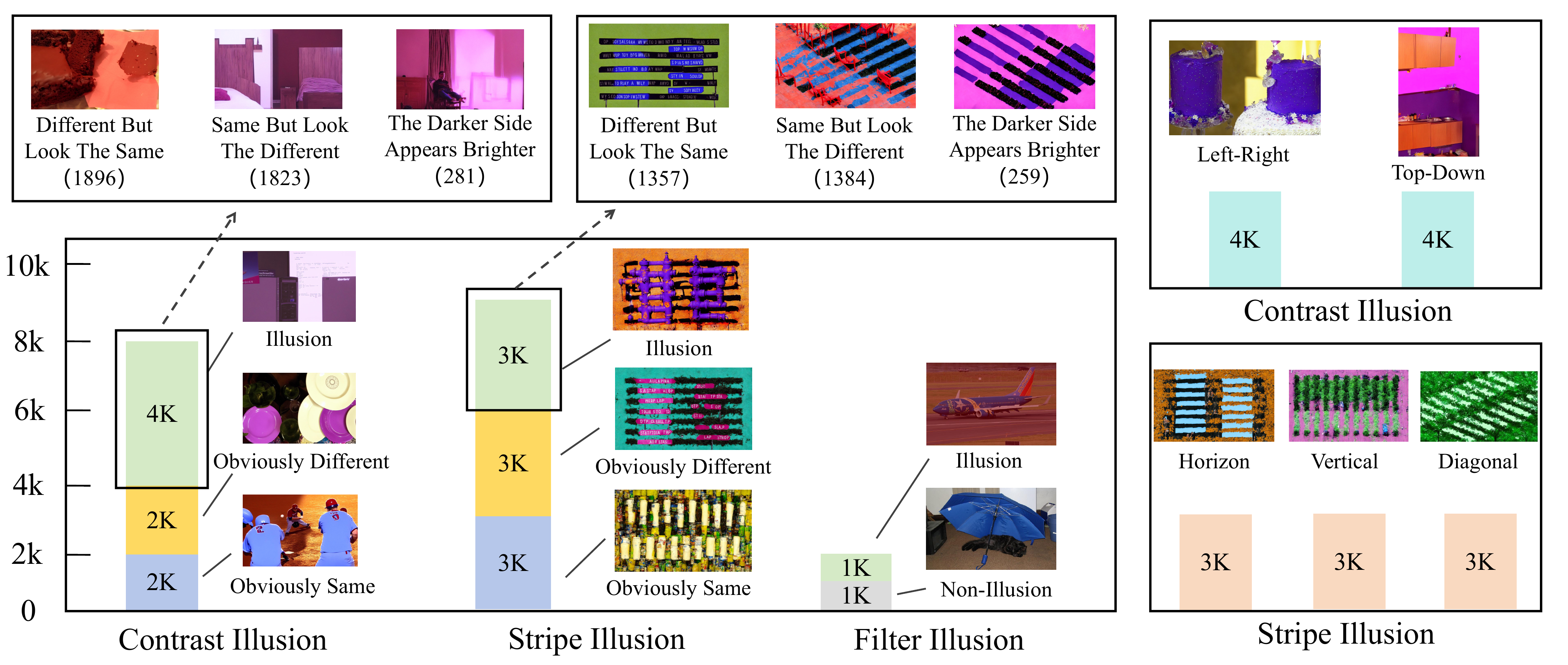}
    \caption{Data statistics of RCID (Realistic Color Illusion Dataset).
    % Each type of illusion exhibits distinct structural patterns, with various effects within each illusion category.
    % For instance, some images appear different yet are actually the same, while others look the same but are actually different.
    }
    % \vskip -1em
    \label{fig:RCID}
\end{figure*}

\paragraph{Procedural Generation Stage:}
We use a procedural function \( f \) to generate simple illusion images, denoted as \( S \), which consist of basic shapes and colors, on which we will condition the trained ControlNet model.

For \textit{Contrast Illusions}, we randomly select two base colors from RGB space: \( C_b \) for the background and \( C_f \) for the foreground (squares). For a color \( C = (r, g, b) \), we use the function \( p \) to adjust its brightness by \( p(C, \mu) = (r \cdot \mu, g \cdot \mu, b \cdot \mu) \), where \( \mu \) scales each RGB component. The background is divided into a darker region \( p(C_b, \mu_{b1}) \) and a brighter region \( p(C_b, \mu_{b2}) \), where \( \mu_{b1} < 1 \, \text{and} \, \mu_{b2} > 1 \). Additionally, two foreground squares, colored \( p(C_f, \mu_{f1}) \) and \( p(C_f, \mu_{f2}) \), are symmetrically placed at random positions \( (x_1, y_1) \) and \( (x_2, y_2) \) on the background.

If human perception differs from the actual pixel values, 
% such as when \( \mu_{f1} = \mu_{f2} \) but the squares appear different in brightness, or when \( \mu_{f1} \neq \mu_{f2} \) but they appear similar, 
the image is classified as an illusion; otherwise, it belongs to the control group (no illusion). We observe that a darker background enhances the perceived brightness of the foreground color, with this effect being strongest when \( C_b \) and \( C_f \) are similar and weakening as the contrast increases. Section 4.2.3 validates this conclusion. Based on this principle, we generate theoretical illusion and non-illusion images, which are later adjusted through human validation. To increase image diversity, we apply random noise \( \eta \) to introduce irregularities at the square edges or make the boundaries less smooth. The generated image \( S_i \) for the contrast illusion can be represented as:

\begin{small}
\[
\begin{aligned}
S = f\big( & p(C_b, \mu_{b1}), p(C_b, \mu_{b2}), p(C_{f}, \mu_{f1}), p(C_{f}, \mu_{f2}), \\
            & (x_1, y_1), (x_2, y_2), \eta \big)
\end{aligned}
\]
\end{small}

Similarly, for \textit{Stripe Illusions}, we randomly generate the background color \( C_b \) and the stripe color \( C_s \). The colored stripes alternate with black stripes, and their colors are adjusted to \( p(C_s, \mu_{s1}) \) and \( p(C_s, \mu_{s2}) \). Based on an initial estimate of human perception of \( \mu_{s1} \) and \( \mu_{s2} \), the images are classified as either \textit{illusion} or \textit{non-illusion}. The generated image \( S \) for the stripe illusion is represented as:

\begin{small}
\[
S = f\left( C_b, p(C_s, \mu_{s1}), p(C_s, \mu_{s2}), \theta, N, \eta \right)
\]
\end{small}

\noindent Here the stripe direction \( \theta \) (horizontal, vertical, or diagonal) and the number of stripes \( N \) are randomly selected. Random noise \( \eta \) is applied to create irregularities, such as curved stripe arrangements or slight misalignments. %, enhancing the diversity.

\paragraph{ControlNet Generating Stage:} 
We pair each simple illusion image \( S \) with a randomly selected caption \( T \) from the MS COCO 2017 dataset, which together serve as input to our model. Starting from the quantized image \( x_0 = S \), the model recursively denoises over multiple time steps following a diffusion-based approach, gradually refining the image to produce a realistic output \( I' \) that retains the intended illusion effects. The final image \( I' \) is obtained by summing the contributions from all time steps \( T \):

\begin{small}
\[
I' = \sum_{t=0}^{T} \left( x_t + \epsilon_\theta(x_t, t, C, S) \right)
\]
\end{small}

\subsection{Filter Illusion Image Generation}

To generate filter illusion images, we first select images from MS COCO 2017 that predominantly contain a specific color \( C_{\text{d}} \) (e.g. red, blue, yellow). We then apply a contrasting color filter \(C_{\text{f}}\) which overlays the original image in the HSV (Hue, Saturation, Value) color space to suppress \(C_{\text{d}}\) by shifting the hue \(H\) of filter and ensuring no pixels fall within \(C_{\text{d}}\)'s HSV range. Despite this, the filtered regions may still visually evoke the original color.

% To generate filter illusion images, we first select images from the MS COCO 2017 dataset that predominantly contain a specific color \( C_{\text{d}} \). After identifying these images, we apply a filter using a contrasting color \( C_{\text{f}} \), which overlays the original image. The goal is to create an illusion by ensuring that the filtered image contains none of the original dominant color, while some people may still perceive these regions as retaining the original color.
% This transformation is done in the HSV (Hue, Saturation, Value) color space to precisely control color shifts. We convert each pixel from RGB to HSV, adjusting the hue \( H \) of the filter to fully suppress \( C_{\text{d}} \), ensuring no pixels fall within the HSV range of this target color. Despite this, the filtered areas may still evoke a visual perception of the original dominant color. %, contributing to the illusion.

\subsection{Question Generation}

For each image \( I' \), we use GPT-4o to generate an image-grounded question \( \mathcal{Q} \) that asks human respondents to compare colors between two regions or identify the color of a specific object.
% , incorporating scene descriptions to make it more conversational. 
For example, in Figure~\ref{fig:illusion_examples}a, the generated questions refer to \textit{the walls on the left and right of the baseball player}.
We then prefix the question with two distinct prompts: one focused on "human perception" (e.g., ``Based on human perception'', ``Through human eye'') and the other on ``pixel values'', each associated with a different answer depending on whether the image contains an illusion. For contrast and stripe illusions, answer options are appended to the question. Examples are shown in Figure~\ref{fig:illusion_examples}.

\subsection{Human Validation and Selection}

To ensure that our generated images truly contain illusions, we conduct a survey on Prolific with a total of 241 participants. Each participant is shown 400 images, including both illusion and non-illusion examples.
% Participants can click on the image to toggle between versions with and without letter (A, B) markings for contrast illusion images. 
We use respondents' answers on illusion images as the human perception ground truth.
For every image in our dataset, we assign 5 participants. 
If more than three participants are deceived by the illusion, the image is classified as an \textit{illusion} image. Conversely, if all five participants provide answers consistent with the pixel values, the image is classified as \textit{non-illusion} and added to the control group. 
On average, we pay \$21 USD per hour.

% Finally, we obtain our dataset \( R = \{(I'_{i}, Q_{i}, H_{i}, A_{i}, \gamma_i)\}_{i=1}^M \), where \( Q_{i} \) represents the corresponding question, \( H_{i} \) corresponds to human perception-based answers, \( A_{i} \) represents the pixel-based answers, and \( \gamma_i \) denotes the category of the image, where \( \gamma_i \in \{\text{illusion}, \text{non-illusion}\} \).

\subsection{Dataset Details} 
Realistic Color Illusion Dataset (RCID, Figure~\ref{fig:RCID}) contains 19,000 images, comprising 8,000 contrast, 9,000 stripe, and 2,000 filter images.
We divide our dataset into 9,500 images for training, 4,750 for development, and the remaining 4,750 for the test set. Each illusion type has distinct structures and subcategories.  %More detailed statistics  can be found in Figure~\ref{fig:RCID}. 

\begin{figure*}[ht!]
\centering
\includegraphics[width=0.9\textwidth]{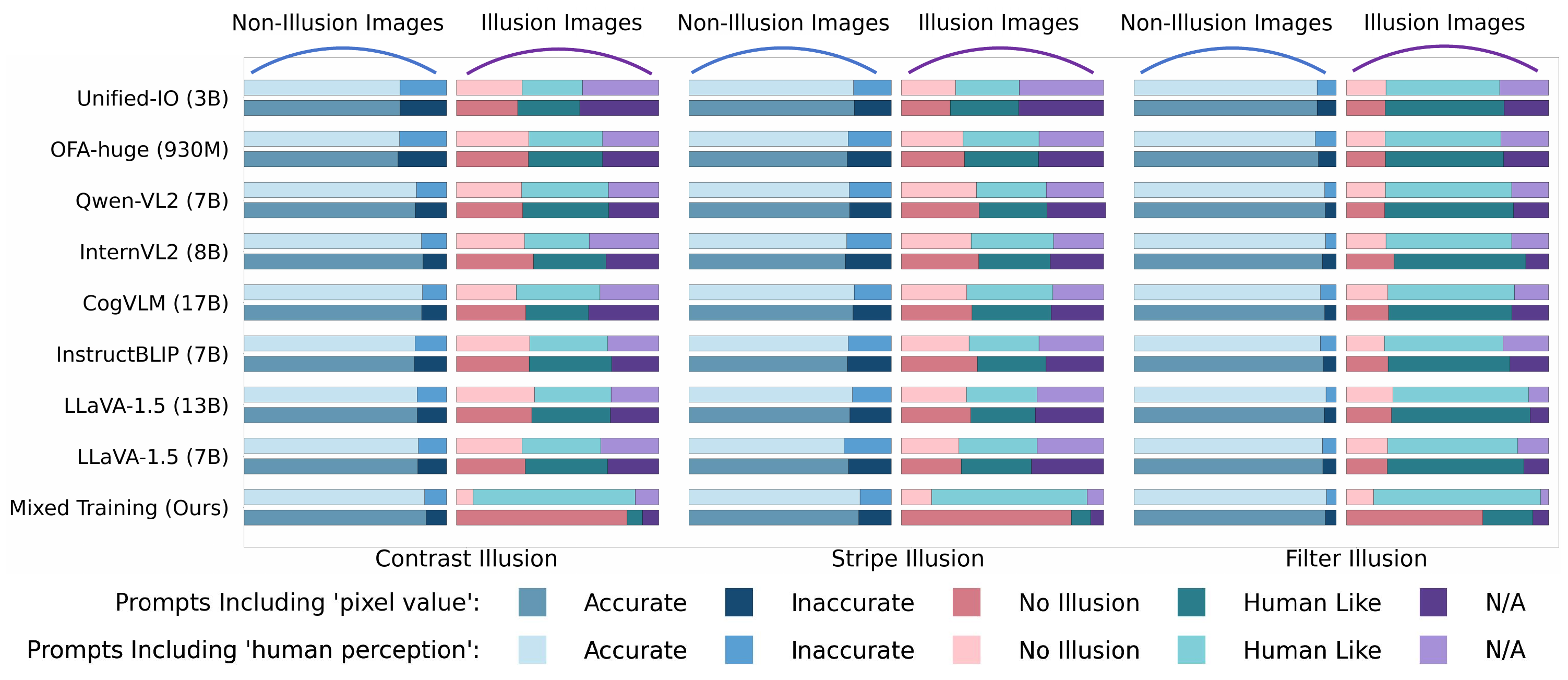}
    \caption{This figure shows the proportion of different model responses across three types of illusions (Contrast, Filter, and Stripe) on our development set. For non-illusion images, we report the proportions of "Accurate" and "Wrong" responses. For illusion images, we categorize responses as "No Illusion" (consistent with pixel values), "Human like," and "N/A." Each image is evaluated using two types of prompts: one based on pixel values and the other based on human perception.}
    % \vskip -1em
    \label{fig:main_performance}
\end{figure*}

\section{Evaluating Vision-Language Models}

\subsection{Experimental Setup}
In this experiment, we evaluate different models' performance on color illusions and explore the impact of external prompts and instruction-tuning on them. We also provide a foundational baseline model based on LLaVA-1.5 (7B) for understanding color illusions.

% \begin{table*}[!ht]
% % \begin{wraptable}{l}{8.3cm}
% % \vskip -1.0em
% \centering
% % \vskip -0.5em
% % \resizebox{0.75\columnwidth}{!}{%
% \resizebox{2.0\columnwidth}{!}{%
% \begin{tabular}{lcccccccccc}
% \toprule
% \multirow{2.5}{*}{Dataset}  & \multicolumn{3}{c}{\bf Contrast Illusion} & & \multicolumn{2}{c}{\bf Filter Illusion} & & \multicolumn{3}{c}{\bf Stripe Illusion}\\
% \cmidrule(lr){2-4} \cmidrule(lr){6-7} \cmidrule(lr){9-11} 
% & Obvious Same & Obvious Diff & Illusion &  & Non-Illusion & Illusion && Obvious Same & Obvious Diff & Illusion  \\
% \midrule
% \# Images & 2000 & 2000 & 4000 && 1000 & 1000 && 3000 & 3000 & 3000 \\
% % \# QA pairs & 4563 & 4782 & 9611 && 2156 & 2031 && 6237 & 6331 & 6124 \\
% \# Image Structure & 2 & 2 & 2 && 1 & 1 && 3 & 3 & 3 \\

% \bottomrule
% \end{tabular}
% }
% \caption{Statistics of the RCID datasets for downstream evaluation on illusion VQA.}
% \label{table:rcid}
% % \vskip -1.0em
% % \end{wraptable}
% \end{table*}

% \begin{figure*}[!ht]
% \centering
%   \includegraphics[width=0.85\textwidth]{imgs/QA-examples.pdf}
%     \caption{Examples of QA pairs from the RCID dataset. For each image, we generate two different prompt versions: one based on pixel values and the other based on human perception.}
%     % \vskip -1em
%     \label{fig:VQA_examples}
% \end{figure*}

\paragraph{Vision-Language Models.} 
We test recent models with our generated illusion dataset to examine whether VLMs exhibit perceptual responses similar to humans. To ensure that their performance on illusion images reflects actual susceptibility to illusions rather than challenges in identifying specified regions and comparing color brightness~\citep{fu2024blinkmultimodallargelanguage, tong2024eyeswideshutexploring}, we focus on open-source models, and fine-tune them on our generated non-illusion simple images and a small set of realistic images to enhance their ability in answering questions about relative colors of different regions in an image. During training, we randomly apply prompts such as 'based on human perception,' 'based on pixel values,' or provide no specific guidance prompt at all. Our models tested include LLaVA~\cite{liu2023visualinstructiontuning}, InstructBLIP~\cite{dai2023instructblipgeneralpurposevisionlanguagemodels}, CogVLM~\cite{wang2024cogvlmvisualexpertpretrained}, InternVL2~\cite{chen2024internvlscalingvisionfoundation}, Qwen-VL2~\cite{bai2023qwenvlversatilevisionlanguagemodel}, OFA~\cite{wang2022ofaunifyingarchitecturestasks} and Unified-IO~\cite{lu2023unifiedio2scalingautoregressive}. 
During inference for contrast and stripe illusions,  questions are presented in a multiple-choice format, while filter illusion questions require open-ended responses about the color of specific objects in the image, as illustrated in Figure~\ref{fig:illusion_examples}.
We also include experiments with closed-source models like GPT-4o and Gemini in the appendix. While their overall color comparison ability is weak, there is still a significant drop in performance on illusion images.

\paragraph{Metrics.}
For images containing illusions, we use the same metrics as in GVIL~\cite{zhang2023grounding}. We track the \textit{Human-like} rate, which measures alignment between human perception and VLM responses by calculating the percentage of cases where the model's responses match human answers. When the prompt explicitly includes "Based on pixel values," this can also be referred to as \textit{Deception Rate}. The \textit{No-Illusion} rate indicates the proportion of examples where the model’s responses align with actual pixel values. If the model's answer aligns with neither the human response nor the pixel values, we categorize those instances as \textit{Not Applicable (N/A)}.
For images without illusions, where human perception matches pixel values, we classify model responses as ``accurate'' or ``inaccurate''.

\begin{figure*}[htbp]
\centering
\includegraphics[width=0.9\textwidth]{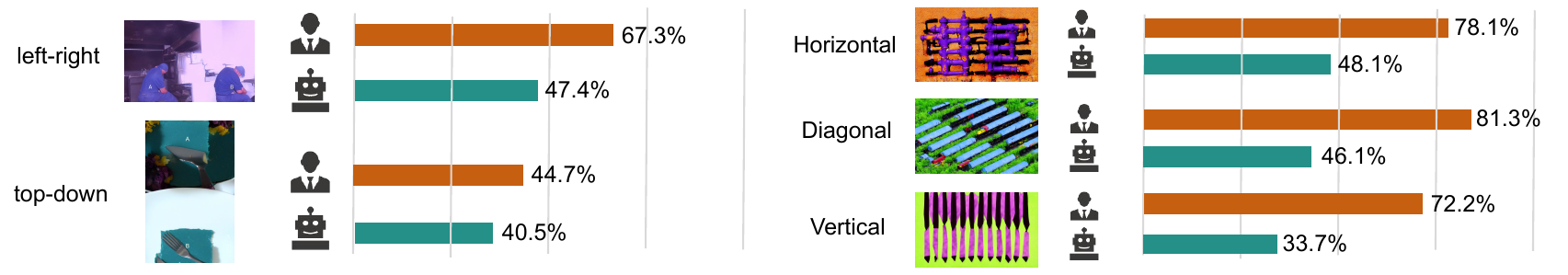}
    \caption{Deception rates of humans and VLMs across different structural patterns.}
    % \vskip -1em
    \label{fig:location}
\end{figure*}

\subsection{Results and Analysis}

We evaluate a range of open-source vision-language models using generated illusion and non-illusion images in our development set, with questions explicitly asking for color judgments 'Based on pixel values' or 'Based on human perception'. The results show that, after fine-tuning on non-illusion images, these models achieve high accuracy (\textgreater 75\%) on non-illusion images, while their accuracy on illusion images is significantly lower.
% , indicating that the models are affected by visual illusions to varying degrees. 
We find that explicitly querying for color judgments 'Based on  pixel values' or 'Based on human perception' does not lead to significant changes in model performance. 
In addition, models are likely to produce responses that are completely inaccurate, matching neither pixel values nor human perception.
This suggests that while the models are misled by color illusions to some extent, they still struggle to fully model human perception. Detailed results can be found in Figure~\ref{fig:main_performance}. 
% We further analyze the factors affecting VLM performance, such as inter-annotator agreement, model size and certain low-level visual features. We also explore whether external prompts or fine-tuning on specific illusion images can improve VLMs' understanding of these illusions.

\paragraph{Inter-Annotator Agreement.}
We refine our final set of illusion images through human validation, retaining only those that deceive at least 3 out of 5 human participants. Before this selection, the average Fleiss' kappa score among the five respondents on the illusion images is 0.648, which increases to 0.806 after filtering. 21.6\% of the selected images deceive all five participants. We use this stricter subset of illusion images as a challenging test set for our models. The results show a slight increase (\textless 5\%) in the proportion of Human-Like responses by the VLMs across all three types of illusions, but overall, the difference remains minimal, indicating that our filtering criterion is robust.

\paragraph{Model Size.}
To investigate the effect of model size, we evaluate the performance of two groups of models, OFA and Unified-IO, on color illusion images. These groups consist of models of varying sizes but share the same overall architecture. We conduct the experiment on contrast illusions, and explicitly emphasize "Based on pixel values" in the questions. 
% As Figure~\ref{fig:model_size_performance} 
Our result shows, with the model size increases, the proportion of human-like responses also increases, while the proportion of responses consistent with actual pixel values decreases. We present OFA's performance in Figure~\ref{fig:model_size_performance} and provide Unified-IO's performance details in the appendix.
% This may be because larger models exhibit greater similarity to human perception.

\begin{figure}[htbp]
\centering
\includegraphics[width=0.8\linewidth]{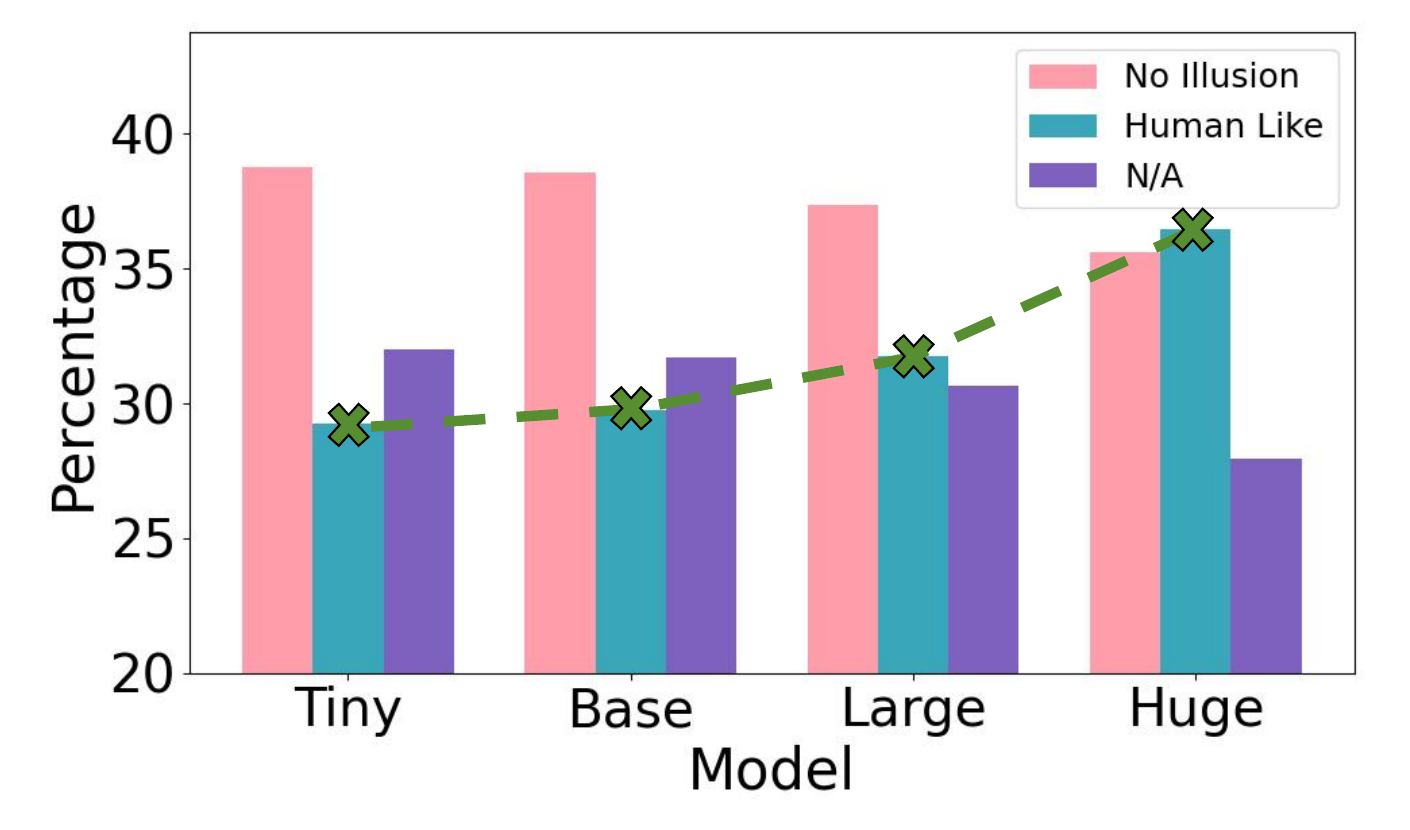}
    \caption{Proportions of 'No Illusion,' 'Human Like,' and 'N/A' responses for OFA models of different sizes on contrast illusion images.}
    % \vskip -1em
    \label{fig:model_size_performance}
\end{figure}

% \begin{figure}[t!]
%     \centering
%     \begin{subfigure}{0.70\linewidth} % Adjusted width to 49% to ensure proper alignment
%         \centering
%         \includegraphics[width=\linewidth]{imgs/OFA-size.pdf}
%         \caption{Performance of OFA models across different sizes.}
%         \label{fig:ofa_performance}
%     \end{subfigure}
%     \hfill
%     \begin{subfigure}{0.70\linewidth} % Adjusted width to 49% to ensure proper alignment
%         \centering
%         \includegraphics[width=\linewidth]{imgs/unified-IO-size.pdf}
%         \caption{Performance of Unified-IO models across different sizes}
%         \label{fig:unified_io_performance}
%     \end{subfigure}
%     \caption{Proportions of 'No Illusion,' 'Human Like,' and 'N/A' responses for models of different sizes on contrast illusion images.}
%     \label{fig:model_size_performance}
% \end{figure}

\paragraph{Illusion design.}
We explore a range of visual factors that may influence the strength of color illusions and compare whether these factors affect human perception and VLMs in the same way. We focus on three potential influencing factors: the orientation of the illusion, the contrast between foreground and background colors (for contrast illusions only), and the number of stripes (for filter illusions only). Overall, our findings indicate that these factors significantly impact the strength of the illusion. 
For example, altering certain factors, such as increasing the color contrast between the foreground and background, can turn a non-illusion image into an illusion image. And these effects are consistent across both humans and VLMs. In these experiments, we use LLaVA-1.5 (7B) as our tested model.

\textit{Illusion orientation.} We randomly generate a set of images with various structural arrangements to analyze how the location of the illusion affects perception in humans and VLMs. For contrast illusions, the images feature left-right and up-down arrangements, while for stripe illusions, they include horizontal, vertical, and diagonal stripes (Figure~\ref{fig:RCID}). We evaluate the error rates of both humans and VLMs across these structures. As illustrated in Figure~\ref{fig:location}, the results show that for contrast illusions, both humans and LLaVA are more frequently deceived by left-right arrangements, while for stripe illusions, they are more often misled by horizontal and diagonal stripes. We find that humans can quickly and accurately perceive color differences between left and right, but are slightly less sensitive to differences between top and bottom.

\begin{table*}[!ht]
\centering
\resizebox{\textwidth}{!}{%
\begin{tabular}{ccccccc}
\toprule
\multirow{2}*{\textbf{Model}} & \multicolumn{2}{c}{\textbf{Contrast Illusion}} & \multicolumn{2}{c}{\textbf{Filter Illusion}} & \multicolumn{2}{c}{\textbf{Stripe Illusion}} \\ \cline{2-7}
              & Pixel-Based Prompts & Human-Based Prompts & Pixel-Based Prompts & Human-Based Prompts & Pixel-Based Prompts & Human-Based Prompts \\ \midrule
LLaVA-1.5 (7B) & 35.4 (41.2) & 32.4 (37.4) & 20.0 (67.8) & 19.8 (64.6) & 30.3 (35.2) & 28.3 (40.1)    \\
Mixed Training (Ours) & 83.9 (\phantom{0}6.9) &  8.1 (79.0) & 68.0 (24.3) & 12.1 (82.3) &  84.5 (\phantom{0}9.8) &  15.4 (76.7)               \\ \bottomrule
\end{tabular}%
}
\caption{The table demonstrates the improvement in the model's understanding of illusions with mixed training. Two prompts are used: "based on pixel value" and "based on human perception." The values represent the proportion of \textit{no-illusion} responses, with parentheses indicating \textit{Human-like} rates.}
\label{table: mix-training}
\end{table*}

\begin{figure}[htbp]
    \centering
    \begin{subfigure}{0.7\linewidth}
        \centering
        \includegraphics[width=\linewidth]{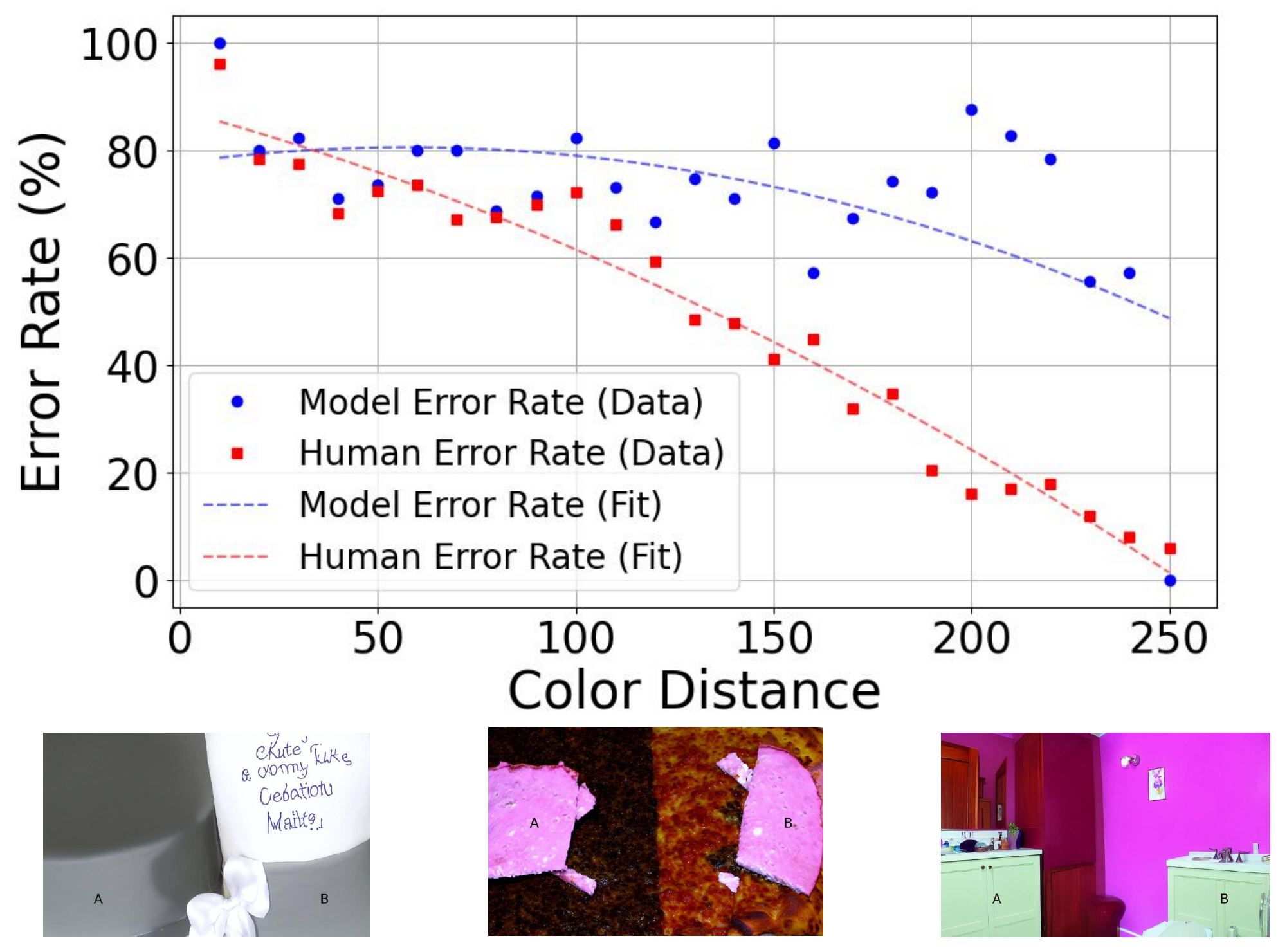}
        \caption{Influence of color differences between background and foreground on intensity of contrast illusion.}
        \label{fig:color-distance}
    \end{subfigure}
    \hfill
    \begin{subfigure}{0.7\linewidth}
        \centering
        \includegraphics[width=\linewidth]{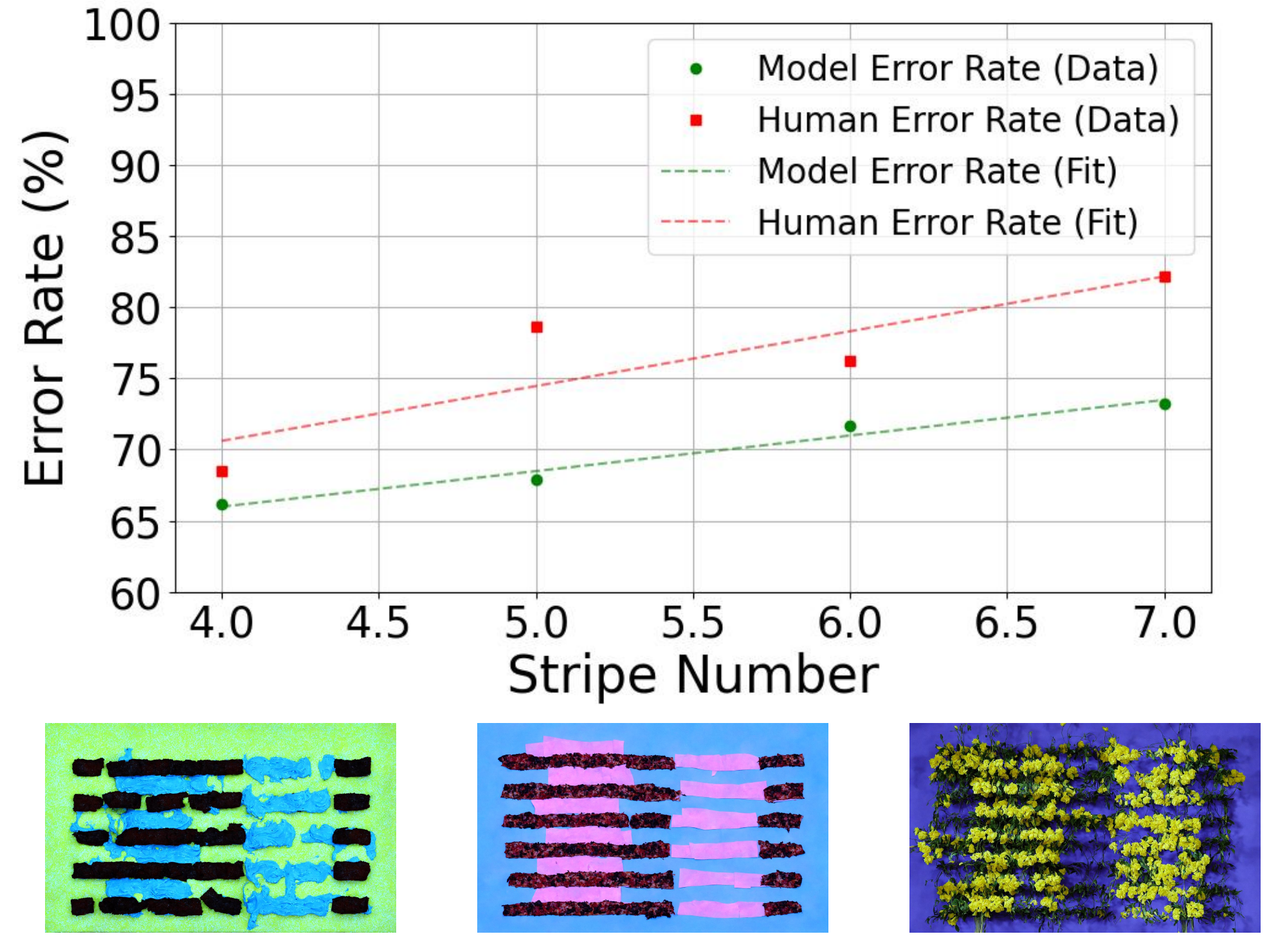}
        \caption{Influence of stripe counts on intensity of stripe illusions.}
        \label{fig:stripe-number}
    \end{subfigure}
    \caption{Error rates of humans and LLaVA across different conditions in color illusions.}
    \label{fig:error_rate}
\end{figure}

\textit{Color distance between the foreground and background.} For contrast illusions, we calculate the correlation between the color distance in RGB space between the foreground and background, and the illusion's intensity.  Figure~\ref{fig:color-distance} shows that as the foreground and background colors become more similar, the contrast illusion becomes most pronounced for both humans and LLaVA. Conversely, when the foreground and background colors are distinctly different, humans and VLMs are better at distinguishing them. 

\textit{Stripe numbers.} For stripe illusions, we examine the effect of the number of stripes on the strength of the stripe illusion, finding that a higher number of stripes intensified the illusion's effect, shown in Figure~\ref{fig:stripe-number}, while fewer stripes diminished the illusion.

\paragraph{Prompting Methods.}\label{sec:prompting}
We explore whether external prompts could alter VLM responses. 
In contrast to our main experiments shown in Figure~\ref{fig:main_performance}, here we aim to identify the model's bias towards predicting answers which are human-like or based on pixel values without specifying this in the question posed to the model.
% In this experiment, we don't explicitly mention "Based on pixel values" or "Based on human perception" to observe whether different prompts would lead VLMs to favor non-illusion answers or human-like responses. 
We consider a number of prompting methods: a simple prompt that only contains questions about color comparisons; chain of thought (CoT), where the model is prompted to first consider if any factors in the image might affect color perception;  emphasizing the illusion by directly informing the model that the image contains an illusion without specifying its type; and providing few-shot examples, using either no-illusion (NI) or human-like (HL) answers as 3-shot examples. Because this experiment uses the LLaVA-7B model, which does not support multi-image input, we combine the few-shot examples and the query image into a single composite image before feeding it into the model~\citep{liu2023visualinstructiontuning}.

\begin{figure}[t!]
\centering
  \includegraphics[width=0.85\linewidth]{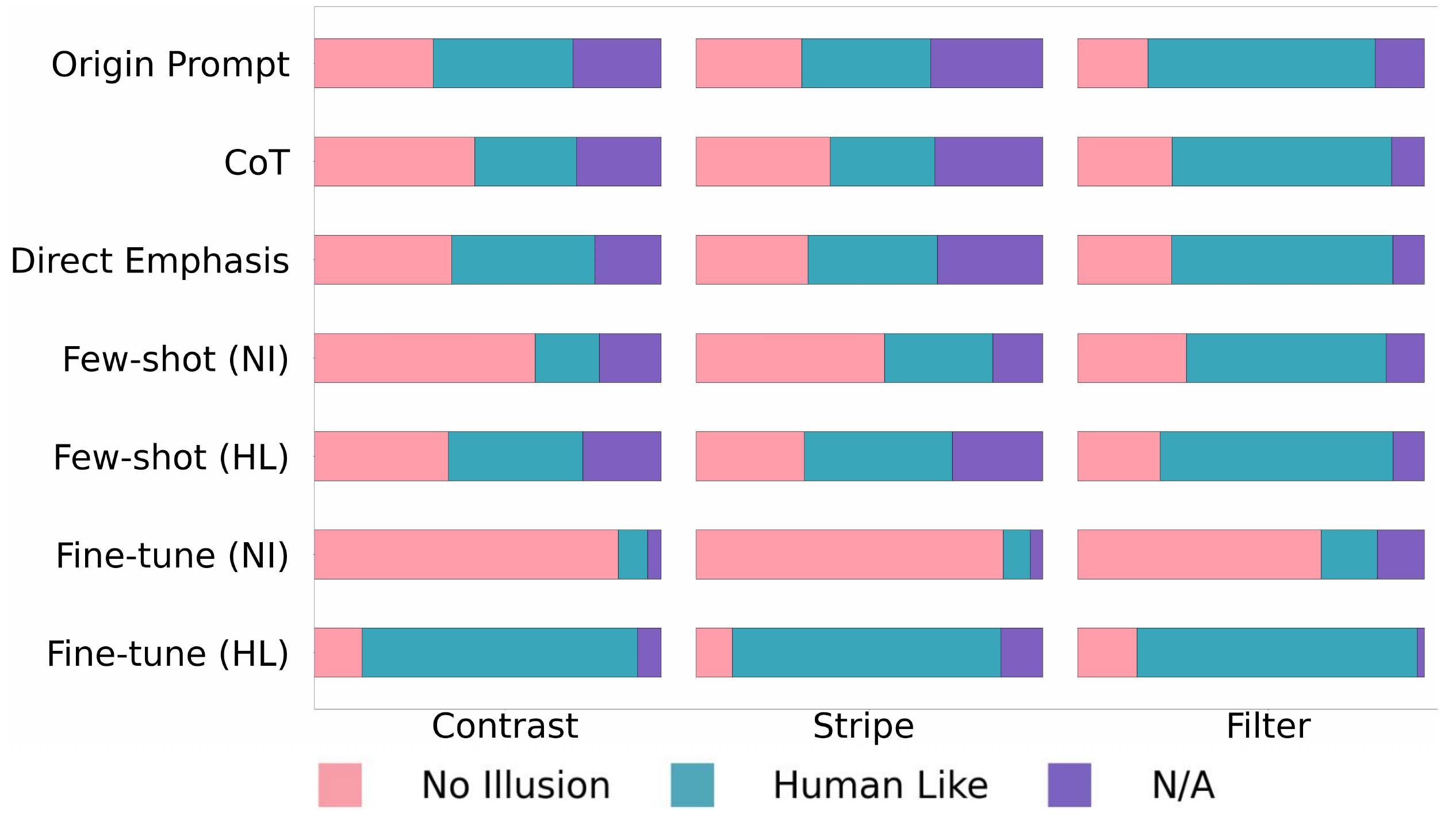}
    \caption{Effect of different prompts and fine-tuning on VLM response preferences across illusion Types.  "NI" denotes datasets using no-illusion examples, while "HL" represents datasets with human-like examples.}
    % \vskip -1em
    \label{fig:prompts_effect}
\end{figure}

Figure~\ref{fig:prompts_effect} show that external prompts have minimal effect on filter illusions. But for contrast and stripe illusions, the effect is more pronounced, particularly when no-illusion answers are used as few-shot examples, increasing the proportion of no-illusion responses by approximately 25\%. However, using human-like responses as few-shot examples does not effectively enhance the model's reproduction of human illusion perception, the proportion of human-like responses even decreases slightly for contrast illusions. Additionally, both direct emphasis and CoT prompts partially increase the proportion of no-illusion responses, although the effect of direct emphasis is relatively weak.

% \begin{table}[htbp]
%     \centering
%     \footnotesize % This reduces the font size
%     \begin{tabular}{p{0.22\linewidth} |>{\centering\arraybackslash}p{0.20\linewidth} >{\centering\arraybackslash}p{0.20\linewidth} >{\centering\arraybackslash}p{0.20\linewidth}}
%         \toprule
%         \textbf{Methods} & \textbf{Contrast} & \textbf{Stripe} & \textbf{Filter} \\
%         \midrule
%         Origin Prompt & 34.27(40.28) & 30.45(37.21) & 20.34(65.57)\\
%         CoT & 46.22(29.41) & 38.75(30.14) & 27.28(63.24) \\
%         Emphasis & 39.61(41.23) & 32.28(37.36) & 27.12(63.77) \\
%         Few-shot (NI) & 63.65(18.49) & 54.35(31.26) & 31.41(57.60) \\
%         Few-shot (HL) & 38.63(38.77) & 31.24(42.65) & 23.86(67.15) \\
%         Fine-tune (NI) & 87.62(8.43) & 88.61(7.83) & 70.23(16.25)\\
%         Fine-tune (HL) & 13.79(79.39) & 10.48(77.46) & 17.13(80.84)\\
%         \bottomrule
%     \end{tabular}
%     \caption{Effect of different prompts and fine-tuning ) on VLM response preferences across illusion Types. Each cell shows the proportion of no-illusion answers (outside parentheses) and human-like answers (inside parentheses). "NI" denotes datasets using no-illusion examples, while "HL" represents datasets with human-like examples.}
%     \label{table: methods}
% \end{table}

\paragraph{Fine-Tuning.}\label{sec:finetuning}
First, we fine-tune two separate LLaVA-7B models on illusion images, with one model trained with responses according to pixel values (NI), and the other trained on human-like responses (HL). The bottom two rows of Figure~\ref{fig:prompts_effect} show a significant effect on the test set, particularly when fine-tuning on no-illusion responses. For both contrast and stripe illusions, fine-tuning on no-illusion responses increases the proportion of no-illusion answers to over 90\%. In contrast, fine-tuning on human-like responses has a weaker effect, suggesting that even with human-like responses, the model struggles to fully grasp human perception in the context of illusion images.

Next, we experiment with a mixed training approach that combines both types of responses. In this setup, we train on question-answer pairs for both types of responses, where questions asking   "Based on pixel values" are paired with pixel-value responses, those asking "Based on human perception" are paired with a human-like response. Table~\ref{table: mix-training} shows the influence of this mixed training on the test set, compared to the base model. These results confirm that training model can effectively distinguish between pixel-based and human perception cues, yielding responses aligned with the user's query.

\vspace{-0.1cm}

\section{Why Color Illusions Affect VLMs}

\subsection{Perceptual Biases from the Visual System}\label{sec:visiononly}

\begin{figure}[t!]
    \centering
    \begin{subfigure}{0.48\linewidth}
        \centering
        \includegraphics[width=\linewidth]{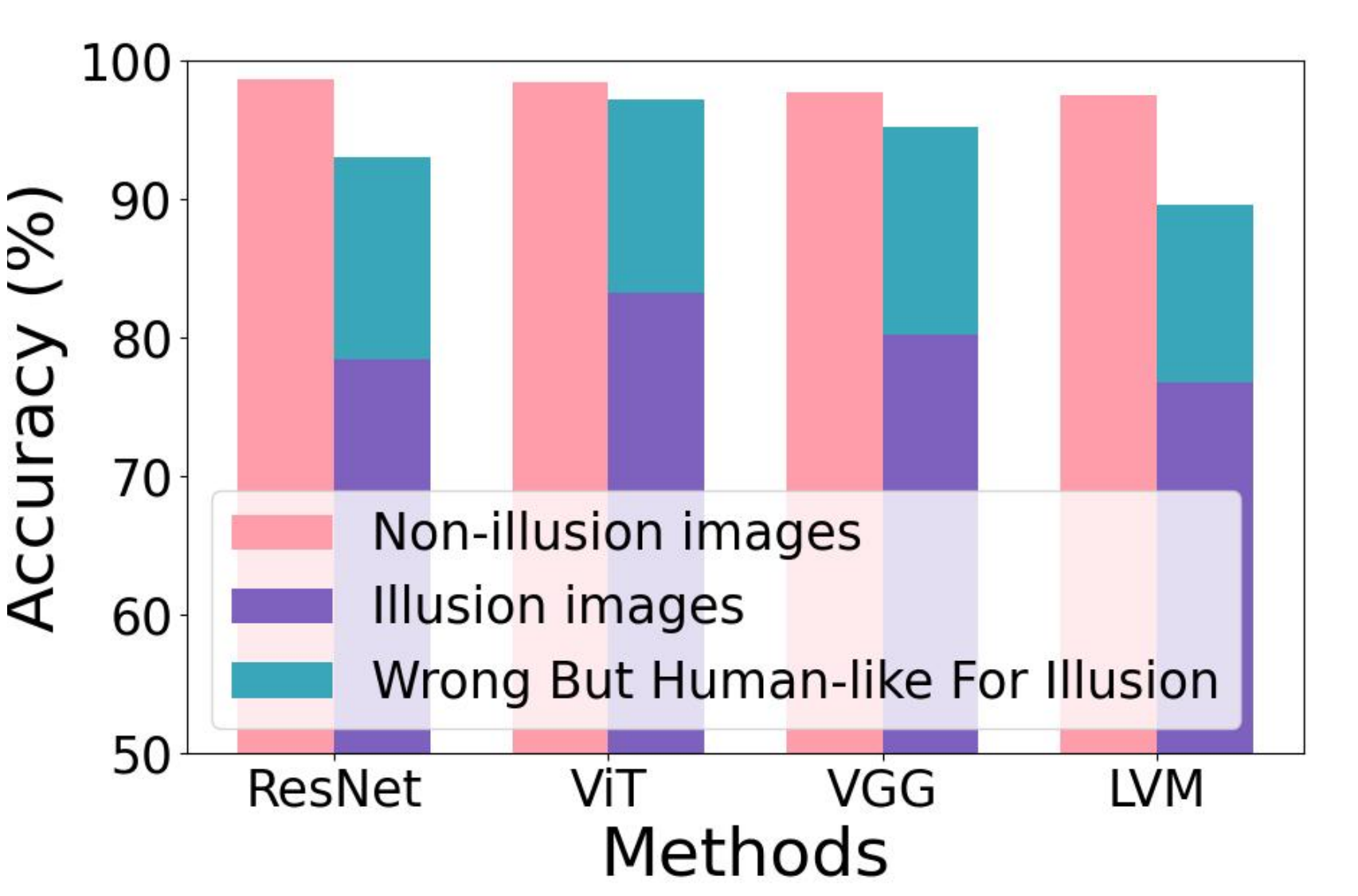}
        \caption{Performance of purely vision models on non-illusion and illusion images.}
        \label{fig:pure_vision_model_performance}
    \end{subfigure}
    \hfill
    \begin{subfigure}{0.48\linewidth}
        \centering
        \includegraphics[width=\linewidth]{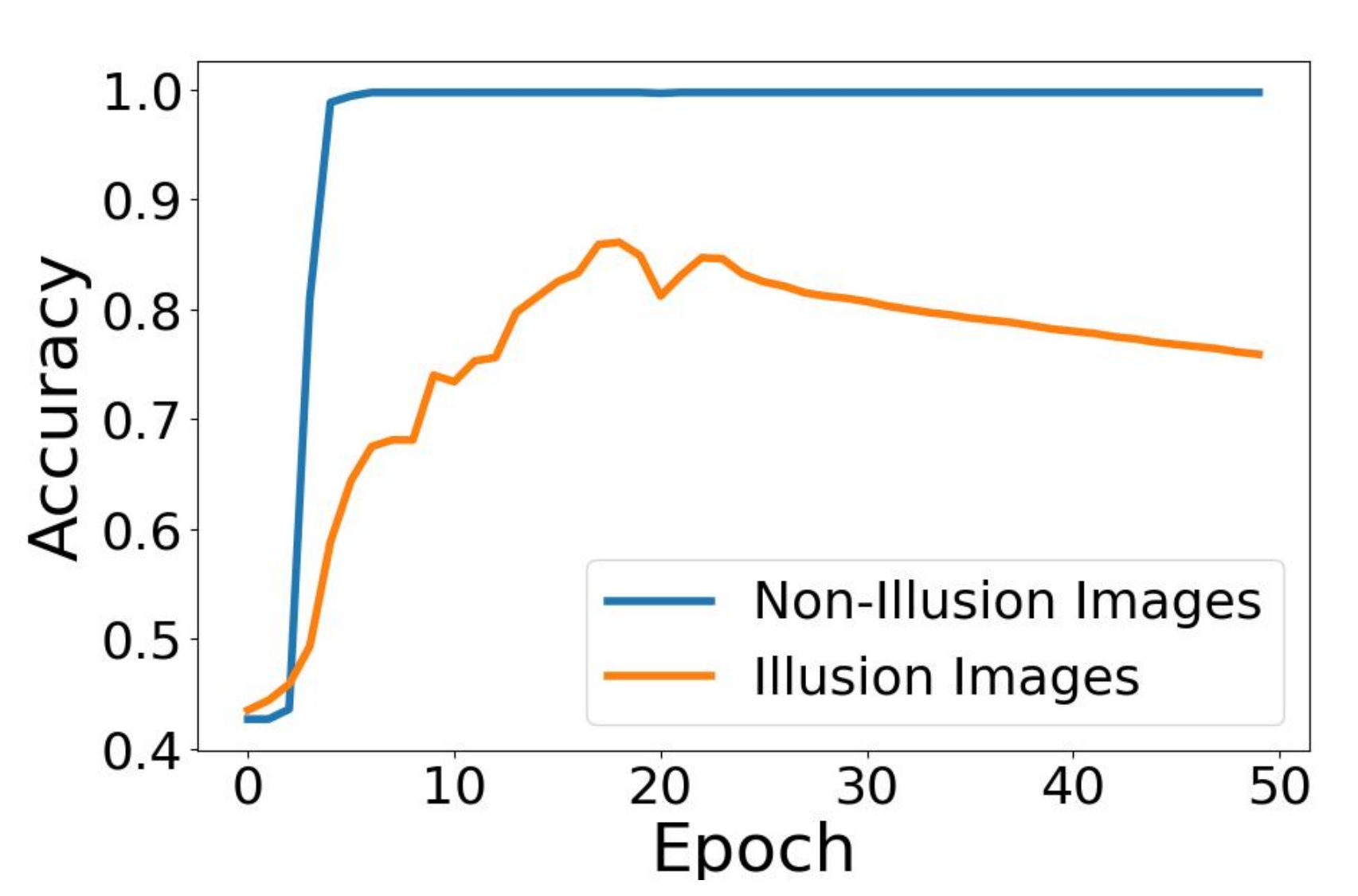}
        \caption{Accuracy of ViT on non-illusion and illusion images over each training epochs.}
        \label{fig:vit_case}
    \end{subfigure}
    \caption{Perceptual biases in purely vision models.}
    \label{fig:error_rate}
\end{figure}

To investigate whether VLMs' perceptual biases originate from their visual components, we test a range of purely visual models, including ResNet~\citep{he2015deepresiduallearningimage}, ViT~\citep{dosovitskiy2021imageworth16x16words}, VGG~\citep{simonyan2015deepconvolutionalnetworkslargescale}, MobileNet~\citep{howard2017mobilenetsefficientconvolutionalneural} and Large Vision Model (LVM)~\citep{bai2024sequential}. We design a simple classification task based on simple rectangle-based contrast illusions.
%, with example images shown in Figure~\ref{fig:pure_vision_task}. 
In this task, we generate backgrounds with varying brightness on both sides and place a rectangle on each side. The vision model is trained as a three-way classifier on 6,000 non-illusion images to predict which rectangle appears darker, or if their colors are identical, stopping when the loss shows no significant decrease. It is then tested on 1,000 non-illusion and 1,000 illusion images.  The results in Figure~\ref{fig:pure_vision_model_performance} show that trained visual models achieve over 95\% classification accuracy on non-illusion images from the test set, but their accuracy (the proportion of classifications aligning with the comparison of pixel values) decreases when presented with illusion images. We also observe that during training, accuracy on non-illusion images quickly improves, while accuracy on illusion images fluctuates significantly. We use Vision Transformer (ViT) as an example in Figure~\ref{fig:vit_case}. These results suggests that even purely vision-based models can be influenced by color illusions.

% \begin{figure}[t!]
% \centering
%   \includegraphics[width=\linewidth]{imgs/pure_vision_task.pdf}
%     \caption{Examples of image classification for color perception tasks, where the model is required to predict which square appears darker or if they appear identical. The first three images are non-illusion cases, while the fourth image is an illusion image, where the left square appears darker than the right, but in reality, both squares are identical in color.}
%     % \vskip -1em
%     \label{fig:pure_vision_task}
% \end{figure}

\subsection{Perceptual Biases from Prior Knowledge}

\begin{figure}[!h]
\centering
  \includegraphics[width=\linewidth]{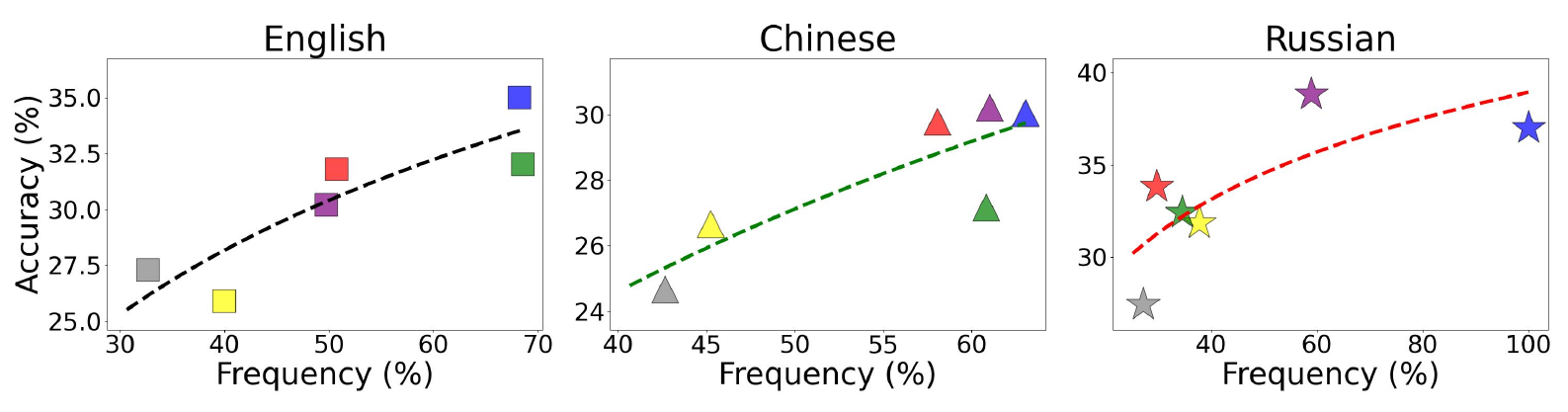}
    \caption{Comparison of color term accuracy and frequency across languages for the PALO model. Each shape and color represents a specific color term (e.g., blue, red, gray).}
    % \vskip -1em
    \label{fig:different_language}
\end{figure}

\paragraph{Influence of Language.} We test the multilingual vision-language model PALO~\citep{PALO} on contrast illusions, posing questions in English, Chinese, and Russian to analyze error rates on images containing target regions across several colors. We hypothesize that colors with more frequent textual modifiers (e.g., \textit{dark blue}, \textit{sky blue}) lead to responses aligned with pixel values rather than human perception. To validate this, we first analyze color descriptor diversity in the training data. For example, in English, 50.7\% of \textit{red} instances include modifiers like \textit{dark red} or \textit{bright red}, while 49.3\% use \textit{red} alone. Figure~\ref{fig:different_language} shows the correlation between color descriptor diversity and pixel-value accuracy for English, Chinese, and Russian.\footnote{In Russian, blue has 100\% descriptor diversity due to distinct terms for \textit{light blue} (\textit{goluboj}) and \textit{deep blue} (\textit{sinij}).} We find that richer descriptions improve model sensitivity to that color, reducing susceptibility to color illusions  in that color range.

% \begin{figure}[htbp]
%     \centering
%     \begin{subfigure}{0.48\textwidth}
%         \centering
%         \includegraphics[width=\linewidth]{imgs/color_term.pdf}
%         \caption{Error rates of Humans and VLMs with varying color differences in contrast illusion images.}
%         \label{fig:color_term}
%     \end{subfigure}
%     \hfill
%     \begin{subfigure}{0.48\textwidth}
%         \centering
%         \includegraphics[width=\linewidth]{imgs/color_term.pdf}
%         \caption{Error rates of Humans and VLMs with different stripe counts in stripe illusion images.}
%         \label{fig:stripe-number}
%     \end{subfigure}
%     \caption{Error rates of Humans and VLMs across different conditions in visual illusions.}
%     \label{fig:color_term}
% \end{figure}

\begin{figure}[t!]
\centering
  \includegraphics[width=0.85\linewidth]{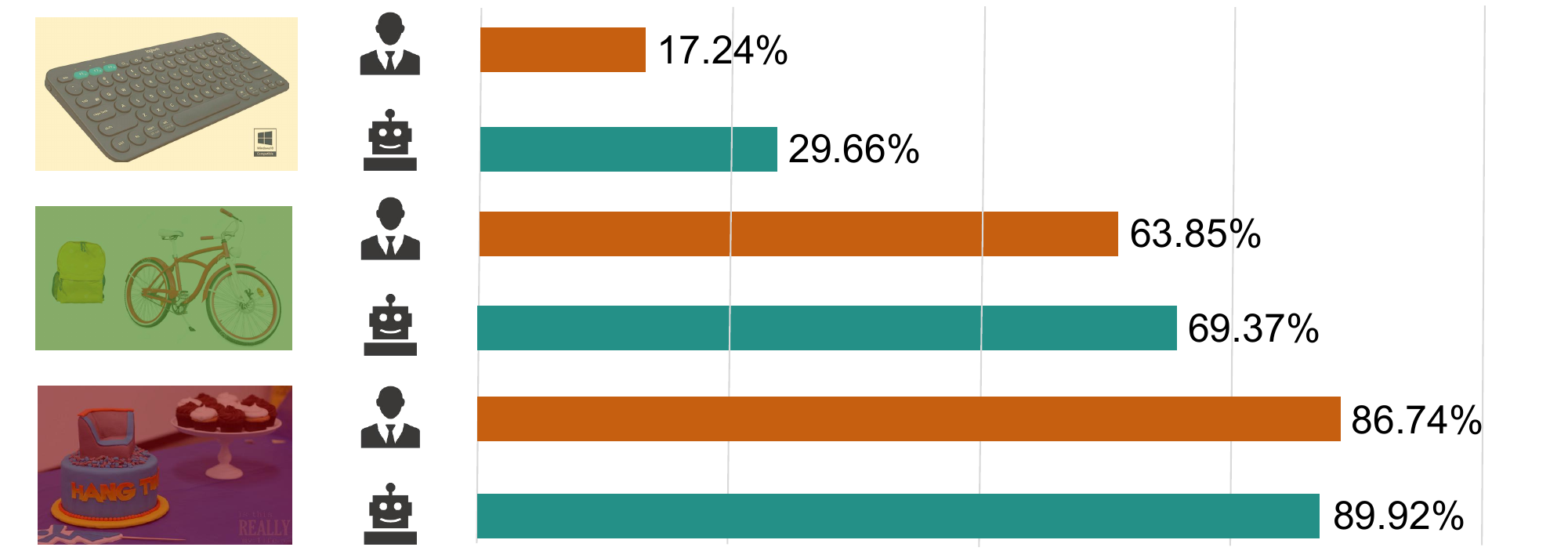}
    \caption{Deception rates of LLaVA and human participants across different types of filter illusions.}
    % \vskip -1em
    \label{fig:filter_illusion_group}
\end{figure}

\paragraph{Influence of Commonsense Knowledge.}
Humans are often deceived by visual illusions due to our prior knowledge. For instance, research suggests that in filter illusions, humans perceive colors that aren’t actually present due to a phenomenon called \textit{chromatic adaptation}~\citep{MacAdam:56}. This allows us to use surrounding context to infer the presence of a filter, mentally "removing" it to perceive the original color. One example is that we always recognize a piece of white paper as white, regardless of the lighting conditions.

To examine whether commonsense knowledge similarly influences VLMs' susceptibility to illusions, we categorized filter illusion images into three groups: single objects, multiple-object compositions, and complex real-world scenes, then analyzed the deception rates for both humans and VLMs across these groups. On non-illusion images, VLMs perform well across all three categories, suggesting that scene complexity does not affect their basic color recognition ability. However, for illusion images, our results in Figure~\ref{fig:filter_illusion_group} show that both humans and VLMs are less likely to be deceived by single-object images, while complex scenes lead to the highest deception rates, underscoring the role of prior knowledge in visual illusions. We assume this is because in complex scenes, there are more objects for models to reference, aiding the inference of colors.

\section{Conclusion}

Our experiments show that current VLMs, influenced by their visual systems and prior human knowledge, can exhibit visual illusions similar to humans. This raises a interesting question: as models become more human-like, might they also inherit human perceptual biases? In fileds like biomedicine~\citep{cui2024biomedical}, such biases could result in flawed visual judgments, while in human-centric tasks like image generation, understanding human perception can benefit. Therefore, VLM behavior should be tailored to application needs, with careful consideration of human-like perceptual biases.

%%%%%%%%% REFERENCES
{\small
\bibliographystyle{ieee_fullname}
\bibliography{egbib}
}

\appendix
\clearpage
\setcounter{page}{1}
\maketitlesupplementary

\section{Data}
\subsection{Prompts for Question Generation}
\begin{figure}[htbp]
\centering
\includegraphics[width=\linewidth]{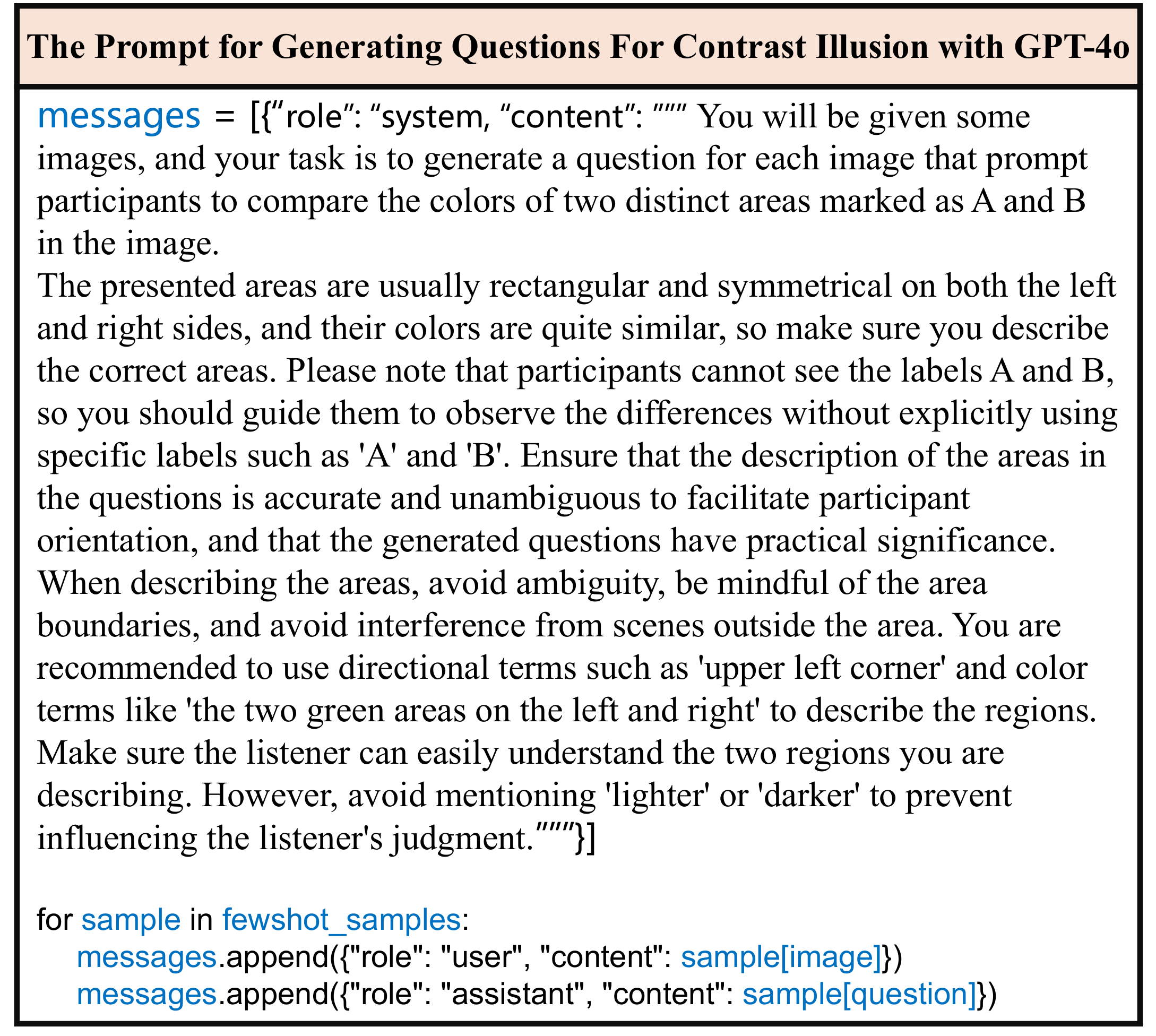}
    \caption{Prompt design for generating color comparison questions on contrast illusions using GPT-4o.}
    % \vskip -1em
    \label{fig:prompt_contrast}
\end{figure}

\begin{figure}[htbp]
\centering
\includegraphics[width=\linewidth]{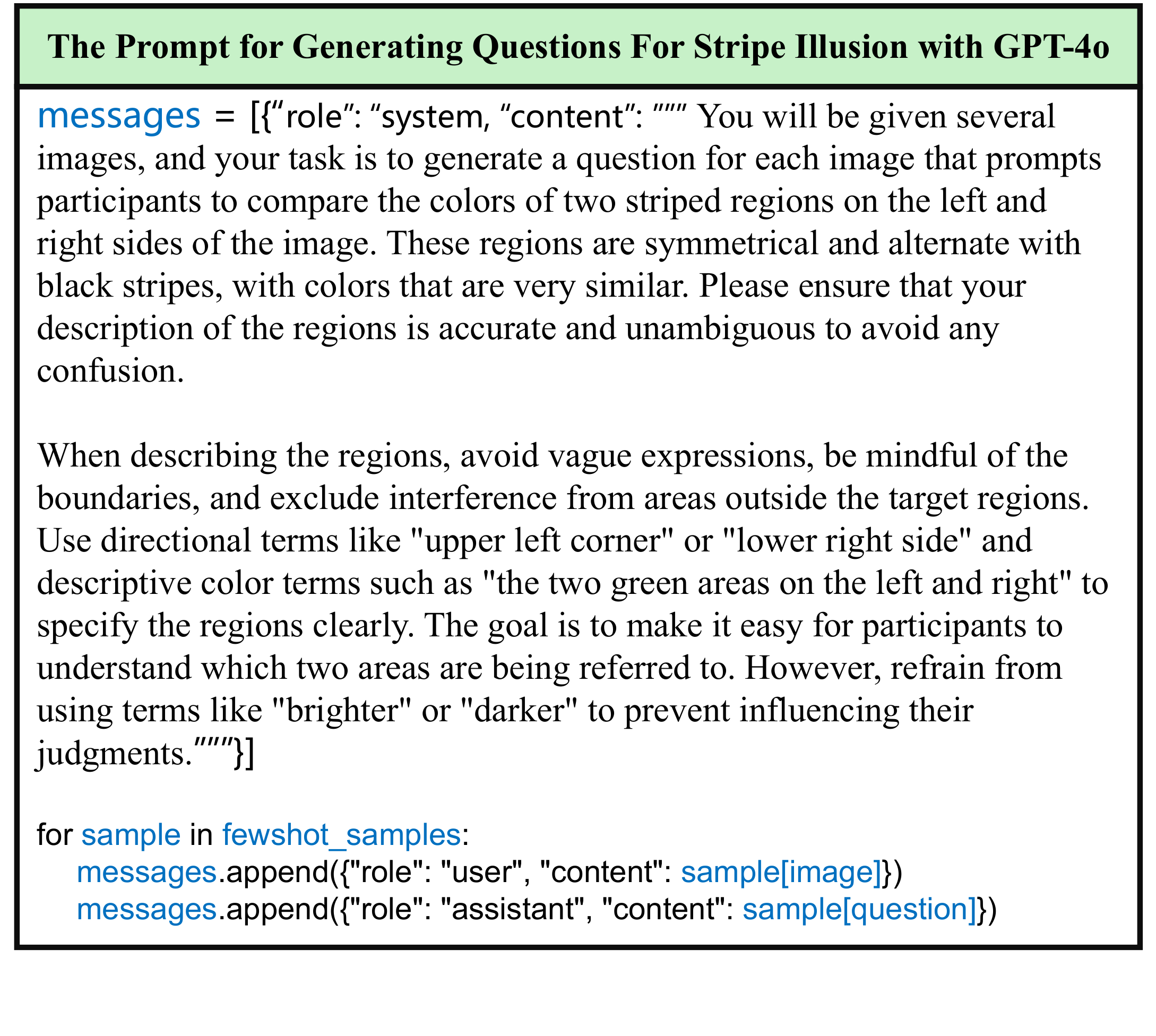}
    \caption{Prompt design for generating color comparison questions on stripe illusions using GPT-4o.}
    % \vskip -1em
    \label{fig:prompt_stripe}
\end{figure}

\begin{figure}[htbp]
\centering
\includegraphics[width=\linewidth]{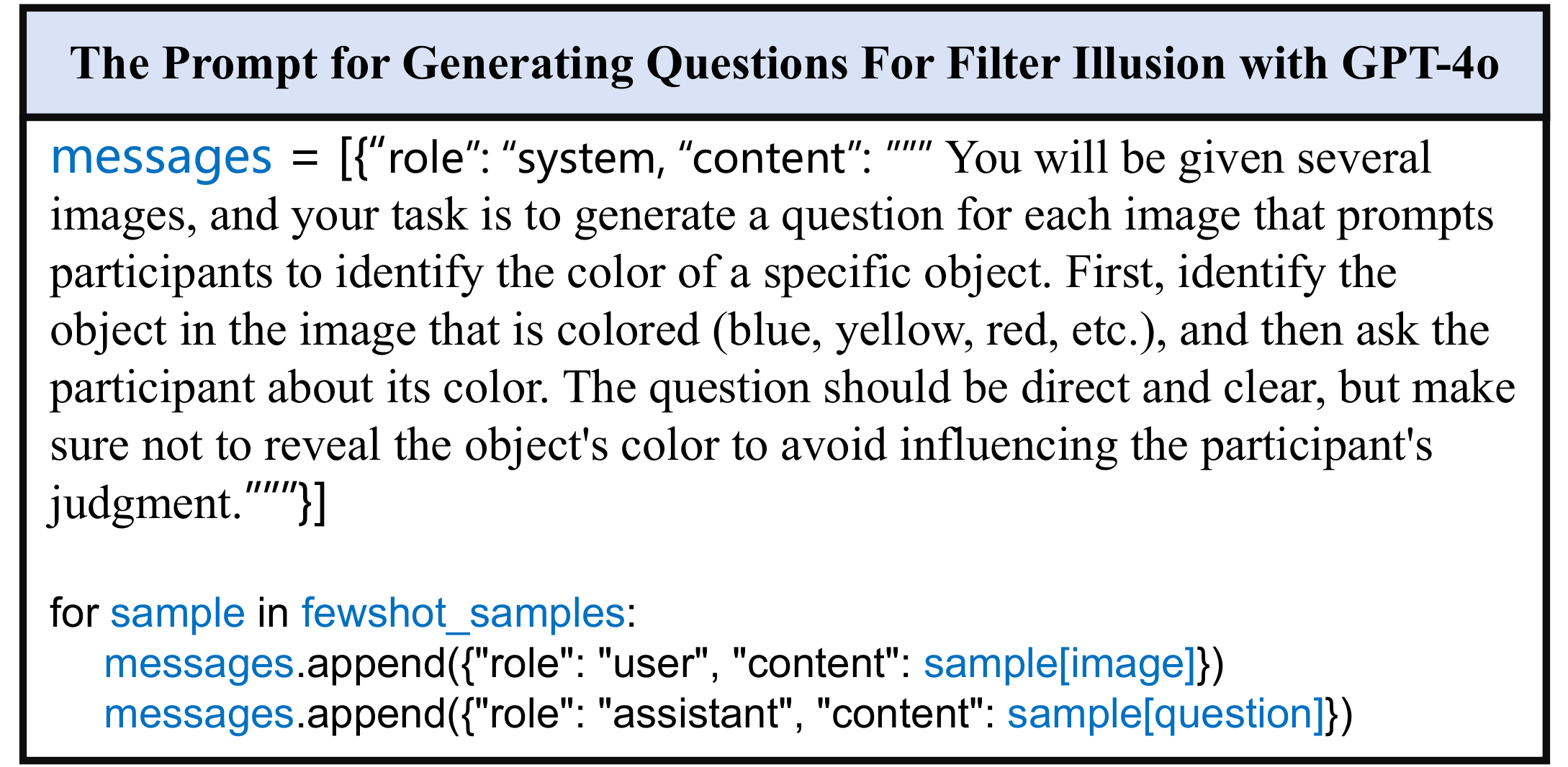}
    \caption{Prompt design for generating color recognition questions on filter illusions using GPT-4o.}
    % \vskip -1em
    \label{fig:prompt_filter}
\end{figure}

Figures~\ref{fig:prompt_contrast},~\ref{fig:prompt_stripe}, and~\ref{fig:prompt_filter} contain the prompts for generating questions for our three different illusion types.

\subsection{Human Data Collection}
\begin{figure*}[t]
\centering
\includegraphics[width=0.95\textwidth]{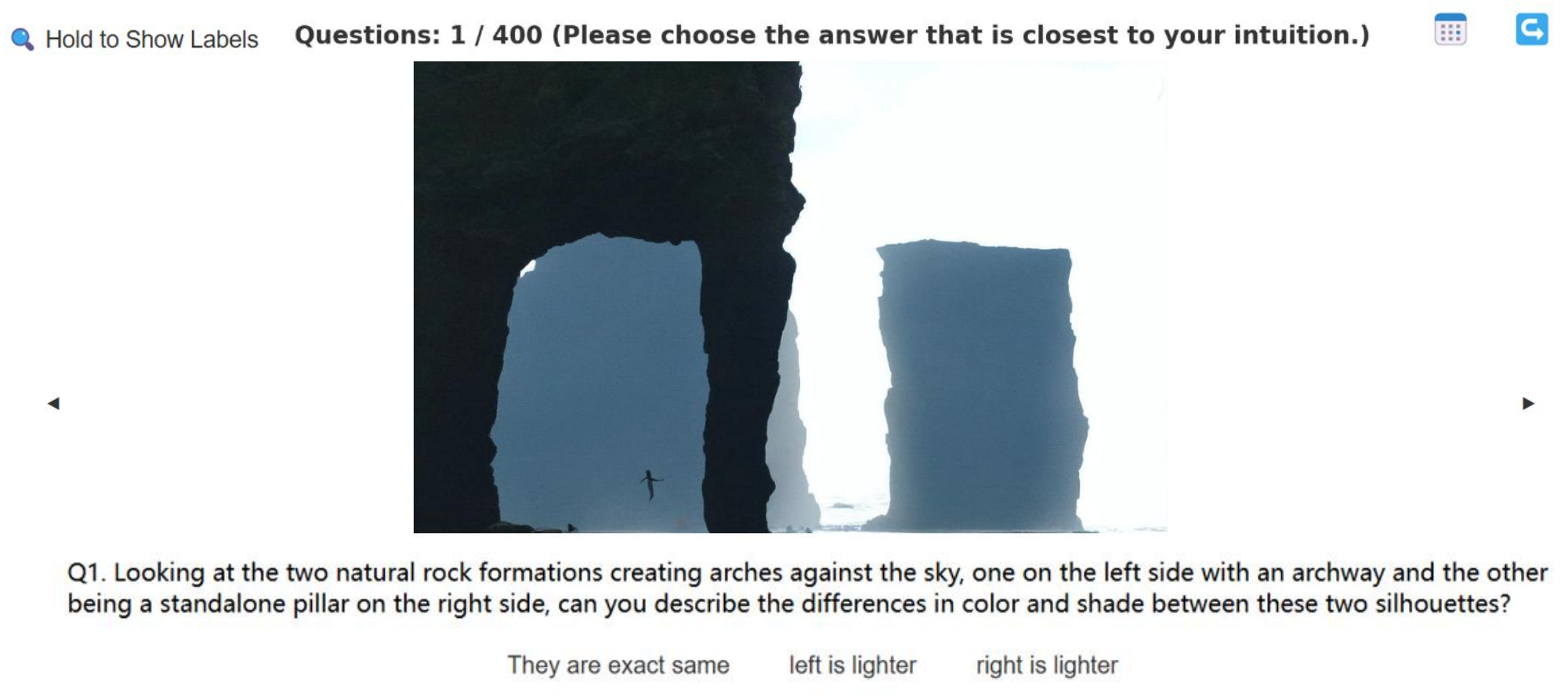}
    \caption{Website interface designed for collecting human responses to color illusion perception tasks.}
    % \vskip -1em
    \label{fig:website}
\end{figure*}

We use Prolific to acquire illusion judgments. Our annotation interface is shown in Figure~\ref{fig:website}. Before annotators begin the task, we provide recommendations for display screen settings (e.g. scaling size, brightness, resolution) and encourage participants to answer in a relatively dark environment to minimize the impact of external lighting. For contrast illusions, participants can toggle between labeled (i.e., with target regions identified with `A' and `B') and unlabeled versions of the image by clicking a button. After every 50 questions completed, the system will enforce a half-minute break to prevent visual fatigue.

\section{Experiments}

% 
% \begin{itemize}
% \item The supplementary can back-reference sections of the main paper, for example, we can refer to \cref{sec:intro};
% \item The main paper can forward reference sub-sections within the supplementary explicitly (e.g. referring to a particular experiment); 
% \item When submitted to arXiv, the supplementary will already included at the end of the paper.
% \end{itemize}
% % 
% To split the supplementary pages from the main paper, you can use \href{https://support.apple.com/en-ca/guide/preview/prvw11793/mac#:~:text=Delete%20a%20page%20from%20a,or%20choose%20Edit%20%3E%20Delete).}{Preview (on macOS)}, \href{https://www.adobe.com/acrobat/how-to/delete-pages-from-pdf.html#:~:text=Choose%20%E2%80%9CTools%E2%80%9D%20%3E%20%E2%80%9COrganize,or%20pages%20from%20the%20file.}{Adobe Acrobat} (on all OSs), as well as \href{https://superuser.com/questions/517986/is-it-possible-to-delete-some-pages-of-a-pdf-document}{command line tools}.

\subsection{Fine-tuning Details}

We conduct all fine-tuning and training on an RTX 6000 using a total of three GPUs. During the pre-training phase, which aims to enhance the model's ability to recognize regions and compare colors, we use 6,000 non-illusion simple images along with our generated realistic non-illusion training set (2,000 for contrast, 3,000 for stripe, and 500 for filter), training for 5 epochs. In subsequent experiments investigating the impact of illusion images during instruction tuning, we add 2,000 contrast illusions, 1,500 stripe illusions, and 500 filter illusions to the training set.

\subsection{Impact of Model Size}

\begin{figure}[H]
\centering
\includegraphics[width=0.8\linewidth]{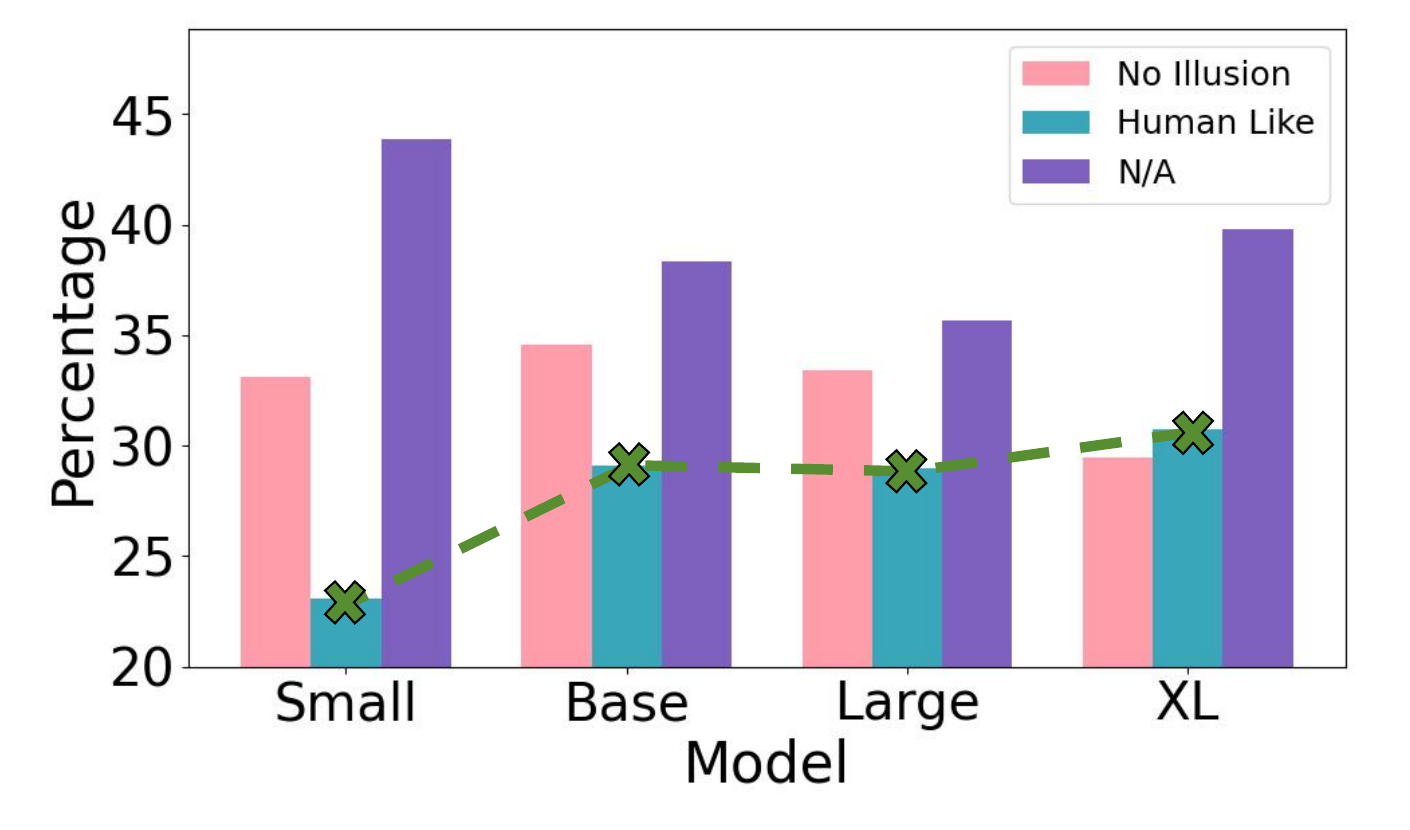}
    \caption{Proportions of `No Illusion,' `Human Like,' and `N/A' responses for Unified-IO models of different sizes on contrast illusion images.}
    % \vskip -1em
    \label{fig:unified-IO-size}
\end{figure}

We additionally provide an example from Unified-IO to demonstrate how model size impacts VLM responses to contrast illusions, shown in Figure~\ref{fig:unified-IO-size}. As the model size increases, the proportion of human-like responses also grows.

\subsection{Impact of Prior Knowledge}

\begin{figure*}[t]
    \centering
    \begin{subfigure}{0.48\textwidth}
        \centering
        \includegraphics[width=\linewidth]{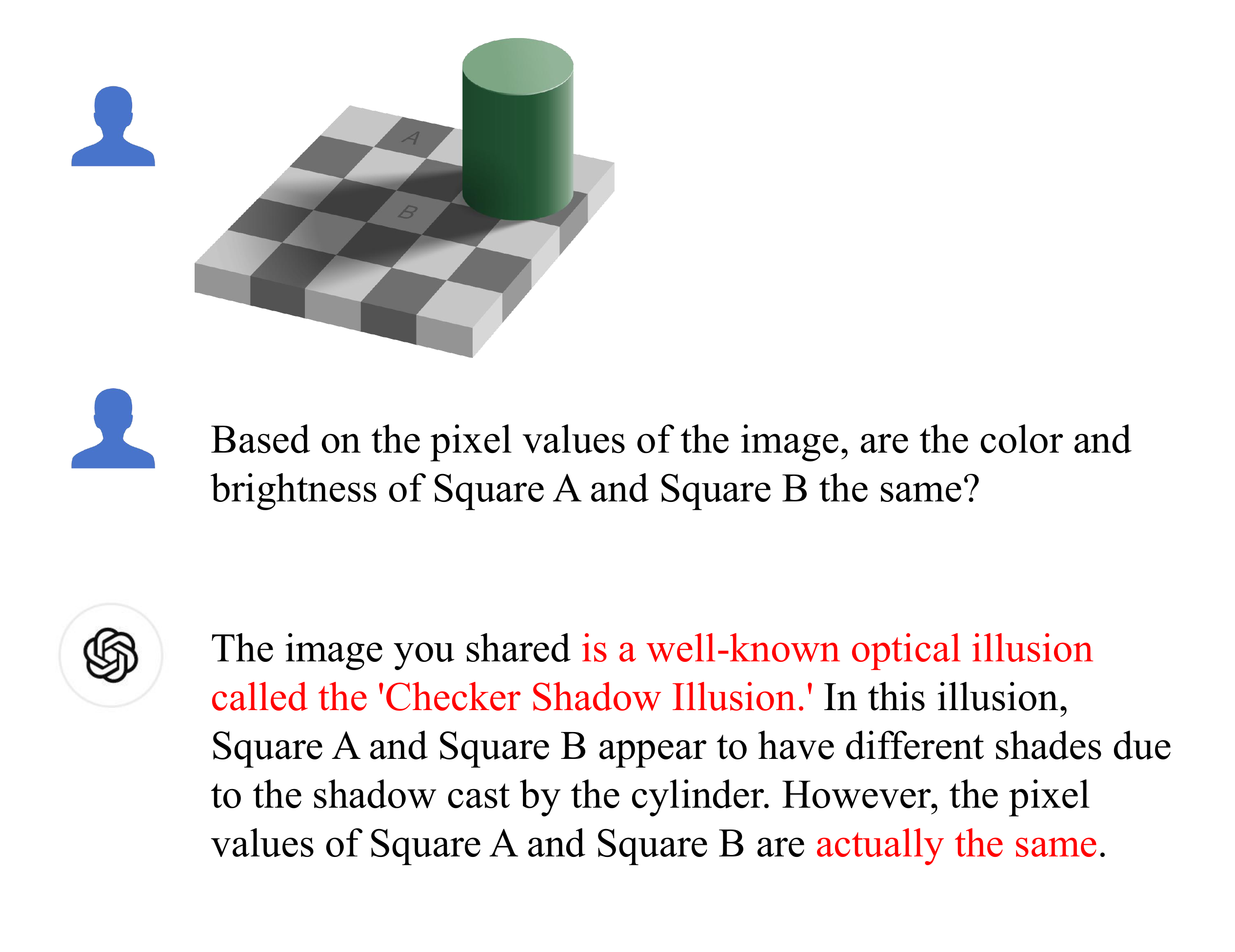}
        \caption{Ask GPT-4 using an website-sourced image}
        \label{fig:ask_using_online_image}
    \end{subfigure}
    \hfill
    \begin{subfigure}{0.48\textwidth}
        \centering
        \includegraphics[width=\linewidth]{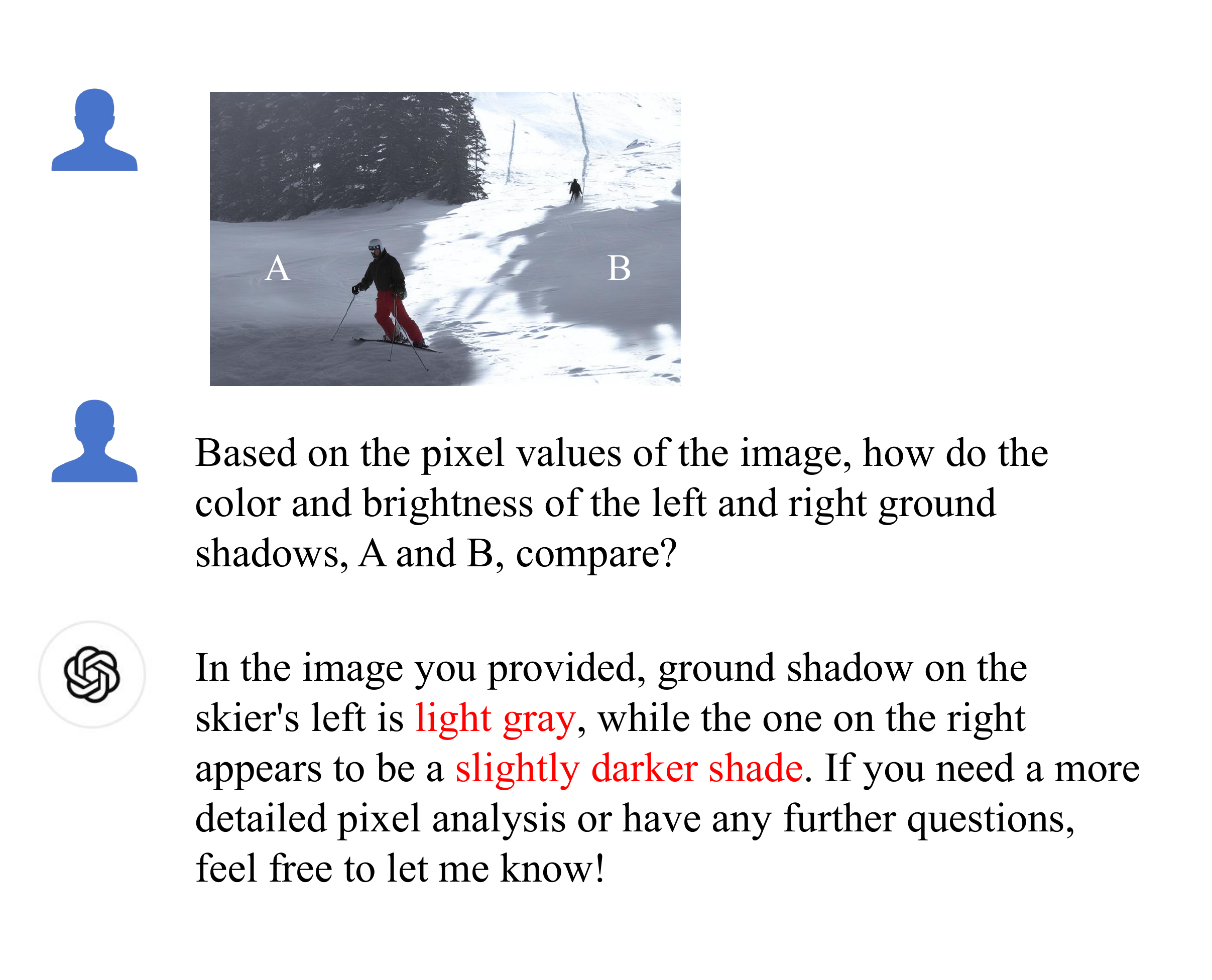}
        \caption{Ask GPT-4 using a generated image}
        \label{fig:ask_using_generated_image}
    \end{subfigure}
    \caption{(a) When presented with a well-known color illusion scraped from the web, GPT-4o can recognize the illusion and identify that the two squares are indeed the same color gray. (b) However, when presenting GPT-4o with an image generated with the same exact shade of gray, it does identify the image as an illusion and is ``deceived''.}
    \label{fig:chat_examples}
\end{figure*}

Figure~\ref{fig:chat_examples} illustrates  how GPT-4o generates different responses for the same gray color block contrast illusion depending on whether the images are sourced from websites or generated. For website-sourced images, GPT-4o often appears to rely on memorized answers rather than deriving responses based on its own observations. To further investigate this behavior, we conduct a statistical analysis of existing illusion datasets, such as IllusionVQA~\citep{shahgir2024illusionvqa}. For each image, we ask GPT-4o to provide an explanation. In 59.3\% of IllusionVQA examples, GPT-4o's responses include phrases like ``well-known'' or ``famous,'' suggesting that these images were likely part of its training data.

\subsection{VLM Performance on Illusion Images Without Fine-tuning for Color QA}

\begin{figure*}[t]
\centering
\includegraphics[width=\textwidth]{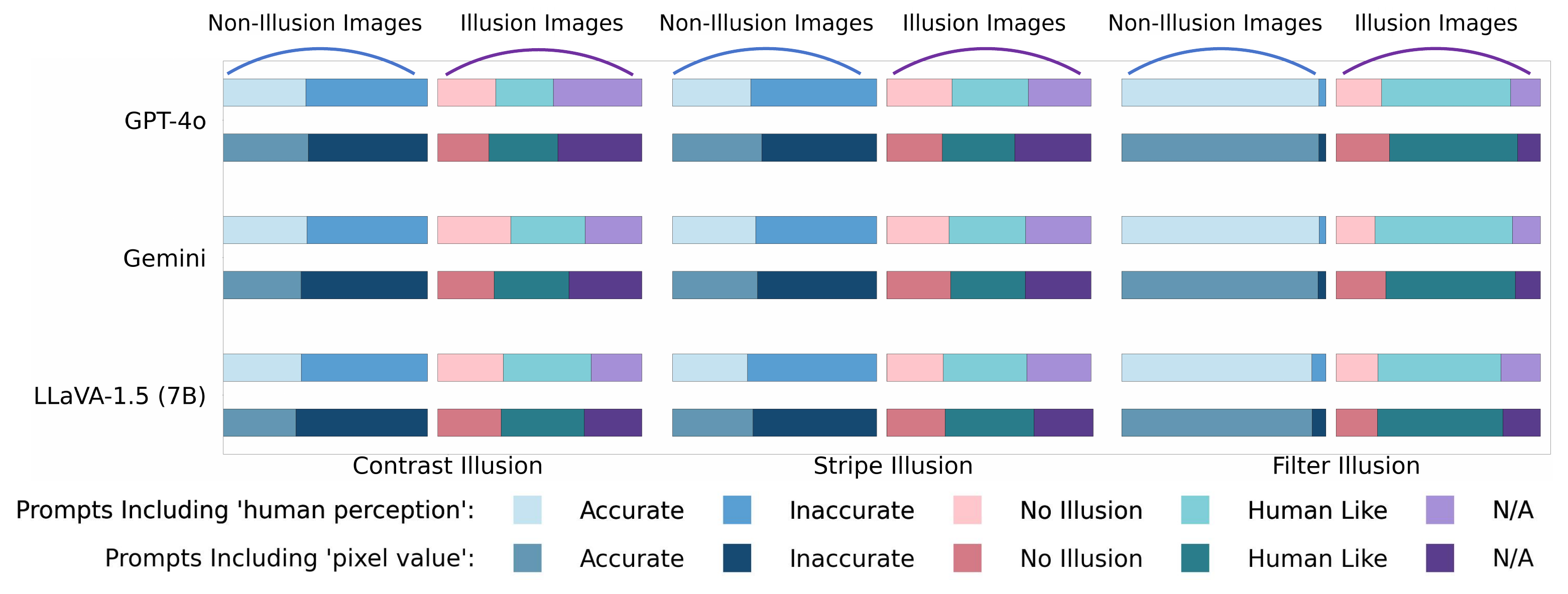}
    \caption{This figure illustrates the proportion of responses from different models across three types of illusions on our development set. The results indicate that VLMs generally perform poorly on tasks involving color comparison (contrast illusion and stripe illusion), while they excel in color recognition tasks (filter illusion). Overall, the responses of VLMs are significantly influenced by color illusions.}
    % \vskip -1em
    \label{fig:main_result_appendix}
\end{figure*}

We evaluate the performance of closed-source models (GPT-4o, Gemini) and the base open-source model (LLaVA-7B) on color illusion images and their corresponding control groups. 
In contrast to our main paper results, none of these models are fine-tuned on the task of answering questions about color comparisons within an image.
The results indicate that, without such fine-tuning, VLMs perform poorly on color depth comparison tasks (contrast illusions and stripe illusions), even on non-illusion images, with accuracy only slightly above random guessing (33.3\%). In contrast, most VLMs perform well on object color recognition tasks (filter illusions). We hypothesize that this may be due to the lack of color comparison data in the training datasets of current VLMs~\citep{fu2024blinkmultimodallargelanguage}. For all three types of color illusions, the proportion of VLM responses consistent with pixel values decreases when tested on illusion images, reflecting a certain degree of visual bias similar to human perception, shown in Figure~\ref{fig:main_result_appendix}.

\subsection{Examples of Color Perception Task For Purely Vision Models}

\begin{figure}[H]
\centering
  \includegraphics[width=\linewidth]{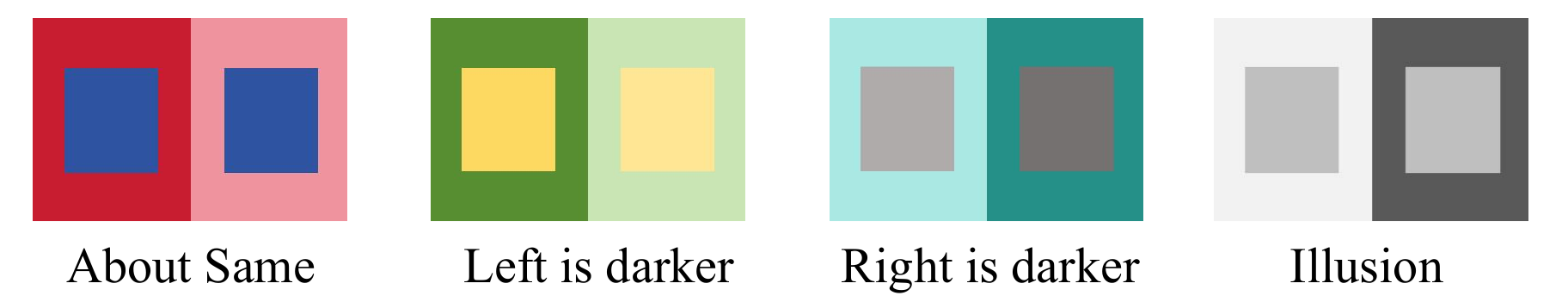}
    \caption{Examples of image classification for color perception tasks, where the model is required to predict which square appears darker or if they appear identical. The first three images are non-illusion cases, while the fourth image is an illusion image, where the left square appears darker than the right, but in reality, both squares are identical in color.}
    % \vskip -1em
    \label{fig:pure_vision_task}
\end{figure}

Figure~\ref{fig:pure_vision_task} shows several examples of the images used in the vision-only Color Perception Task described in Section~\ref{sec:visiononly}.

\end{document}